\documentclass[preprint,3pt]{elsarticle}

\usepackage{hyperref}

\usepackage{amsmath, amssymb}
\usepackage[capitalize]{cleveref}
\usepackage{color,soul}
\usepackage{tabularx}
\usepackage{MnSymbol}
\usepackage[capitalize]{cleveref}
\usepackage{amsfonts}

\usepackage{url}
\usepackage{color,soul}
\usepackage{bm}
\usepackage{multirow}
\usepackage[font=small]{caption}
\usepackage{subcaption}
\usepackage{afterpage}
\usepackage{arydshln}
\usepackage{url}
\usepackage[shortlabels]{enumitem}
\usepackage{bigstrut}
\usepackage{array}
\usepackage{microtype}
\usepackage{pifont}
\usepackage{textcomp}
\usepackage{siunitx}
\usepackage{flushend}
\usepackage{booktabs,dcolumn}

\newcolumntype{d}[1]{D..{#1}}

\DeclareMathOperator*{\argmax}{arg\,max}

\journal{Journal of Elsevier}

\begin{document}

\begin{frontmatter}

\title{Conversational Transfer Learning for Emotion Recognition}

\author[address1]{Devamanyu Hazarika}
\ead{hazarika@comp.nus.edu.sg}

\author[address3]{Soujanya Poria\corref{correspondingauthor}}
\cortext[correspondingauthor]{Corresponding author.}
\ead{soujanya\_poria@sutd.edu.sg}

\author[address1]{Roger Zimmermann}
\ead{rogerz@comp.nus.edu.sg}

\author[address2]{Rada Mihalcea}
\ead{mihalcea@umich.edu}

\address[address1]{School of Computing, National University of Singapore}
\address[address2]{Computer Science \& Engineering, University of Michigan, USA}
\address[address3]{Information Systems Technology and Design, Singapore University of Technology and Design}

\begin{abstract}
Recognizing emotions in conversations is a challenging task due to the presence of contextual dependencies governed by self- and inter-personal influences. Recent approaches have focused on modeling these dependencies primarily via supervised learning. However, purely supervised strategies demand large amounts of annotated data, which is lacking in most of the available corpora in this task. To tackle this challenge, we look at transfer learning approaches as a viable alternative. Given the large amount of available conversational data, we investigate whether generative conversational models can be leveraged to transfer affective knowledge for detecting emotions in context. We propose an approach, \textit{TL-ERC}, where we pre-train a hierarchical dialogue model on multi-turn conversations (source) and then transfer its parameters to a conversational emotion classifier (target). In addition to the popular practice of using pre-trained sentence encoders, our approach also incorporates recurrent parameters that model inter-sentential context across the whole conversation. Based on this idea, we perform several experiments across multiple datasets and find improvement in performance and robustness against limited training data. TL-ERC also achieves better validation performances in significantly fewer epochs. Overall, we infer that knowledge acquired from dialogue generators can indeed help recognize emotions in conversations.
\end{abstract}


\end{frontmatter}


\section{Introduction} \label{sec:intro}

Emotion Recognition in Conversations (ERC) is the task of detecting emotions from utterances in a conversation. It is an important task with applications ranging from dialogue understanding to affective dialogue systems~\cite{poria2019emotion}. Apart from the traditional challenges of dialogue understanding, such as intent-detection, contextual grounding, and others~\cite{chen2017survey}, ERC presents additional challenges as it requires the ability to model emotional dynamics governed by self- and inter-speaker influences at play~\cite{DBLP:conf/naacl/HazarikaPZCMZ18}. Further complications arise due to the limited availability of annotated data --- especially in multimodal ERC --- and the variability in annotations owing to the subjectivity of annotators in interpreting emotions.

In this work, we focus on these issues by investigating a framework of sequential inductive \textit{transfer learning} (TL)~\cite{pan2010survey}. In particular, we attempt to transfer contextual affective information from a generative conversation modeling task to ERC. We name this framework TL-ERC.

\begin{figure}[h]
	\centering
	\includegraphics[width=0.6\linewidth]{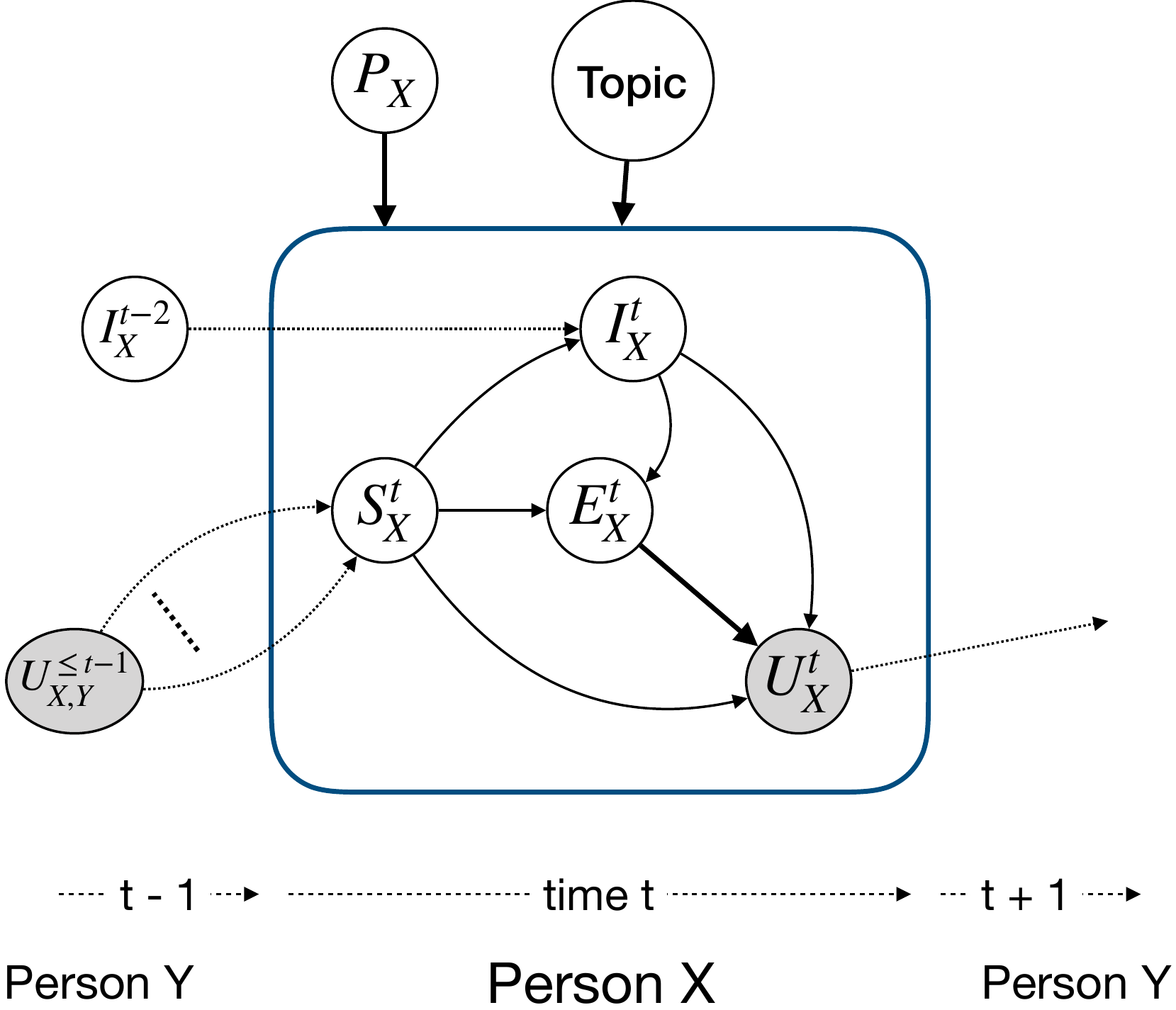}
	\caption{\footnotesize{Dyadic conversation --- between person X and Y --- are governed by interactions between several latent factors. Emotions are a crucial component in this generative process. In the illustration, $P$ represents the personality of the speaker; $S$ represents speaker-state; $I$ denotes the intent of the speaker; $E$ refers to the speaker's emotional state, and $U$ refers to the observed utterance. Speaker personality and the topic always condition upon the variables. At turn $t$, the speaker conceives several pragmatic concepts such as argumentation logic, viewpoint, and inter-personal relationship - which we collectively represent using the speaker-state $S$~\cite{hovy1987generating}. Next, the intent $I$ of the speaker gets formulated based on the current speaker-state and previous intent of the same speaker (at $t-2$). These two factors influence the emotional feeling of the speaker, which finally manifests as the spoken utterance~\cite{poria2019emotion}. }}
	\label{fig:controlling_vars}
\end{figure}

\textit{But why should generative modeling of conversations acquire knowledge on emotional dynamics?} To answer this question, we first observe the role of emotions in conversations. Several works in the literature have indicated that emotional goals and influences act as latent controllers in dialogues~\cite{weigand2017emotions,sidnell2012handbook}. \citet{poria2019emotion} demonstrated the interplay of several factors, such as the topic of the conversation, speakers' personality, argumentation-logic, viewpoint, and intent, which modulate the emotional state of the speaker and finally lead to an utterance.~\cref{fig:controlling_vars} illustrates these dependencies, which elaborate emotional factor as a critical latent state in the overall generative process of dialogues.

\begin{figure*}[t]
	\centering
	\includegraphics[width=\linewidth]{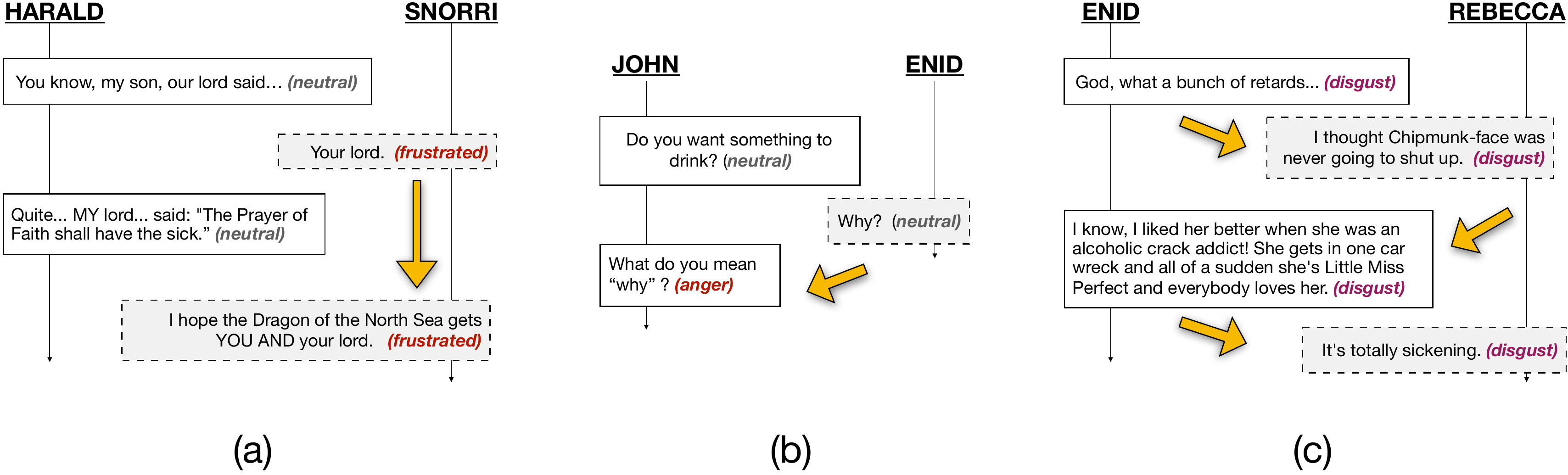}
	\caption{\footnotesize{Samples from \textit{Cornell Movie Dialog Corpus}~\cite{danescu2011chameleons}. The examples demonstrate various kinds of emotional influences, such as \textit{emotional inertia}, \textit{mirroring}, etc., that manifest in natural conversations. }}
	\label{fig:examples}
\end{figure*}

The interactions between these latent factors lead to diverse emotional dynamics within the conversations.~\cref{fig:examples} provides some examples demonstrating such patterns. In the figure, conversation (a) illustrates the presence of \textit{emotional inertia}~\cite{koval2012changing} which occurs though self-influences in emotional states. The character \textit{Snorri} maintains a frustrated emotional state by not being affected/influenced by the other speaker. Whereas, conversation (b) and (c) demonstrate the role of inter-speaker influences in emotional transitions across turns. In (b), the character \textit{John} is triggered for an \textit{emotional shift} due to influences based on his counterpart's responses, while (c) demonstrates the effect of \textit{mirroring}~\cite{navarretta2016mirroring} which often arises due to topical agreement between speakers. All these examples demonstrate the presence of such emotional dynamics that are not just inherent in the conversations but also help shape them up~\cite{poria2019emotion}.

\begin{figure}[t]
	\centering
	\includegraphics[width=.5\linewidth]{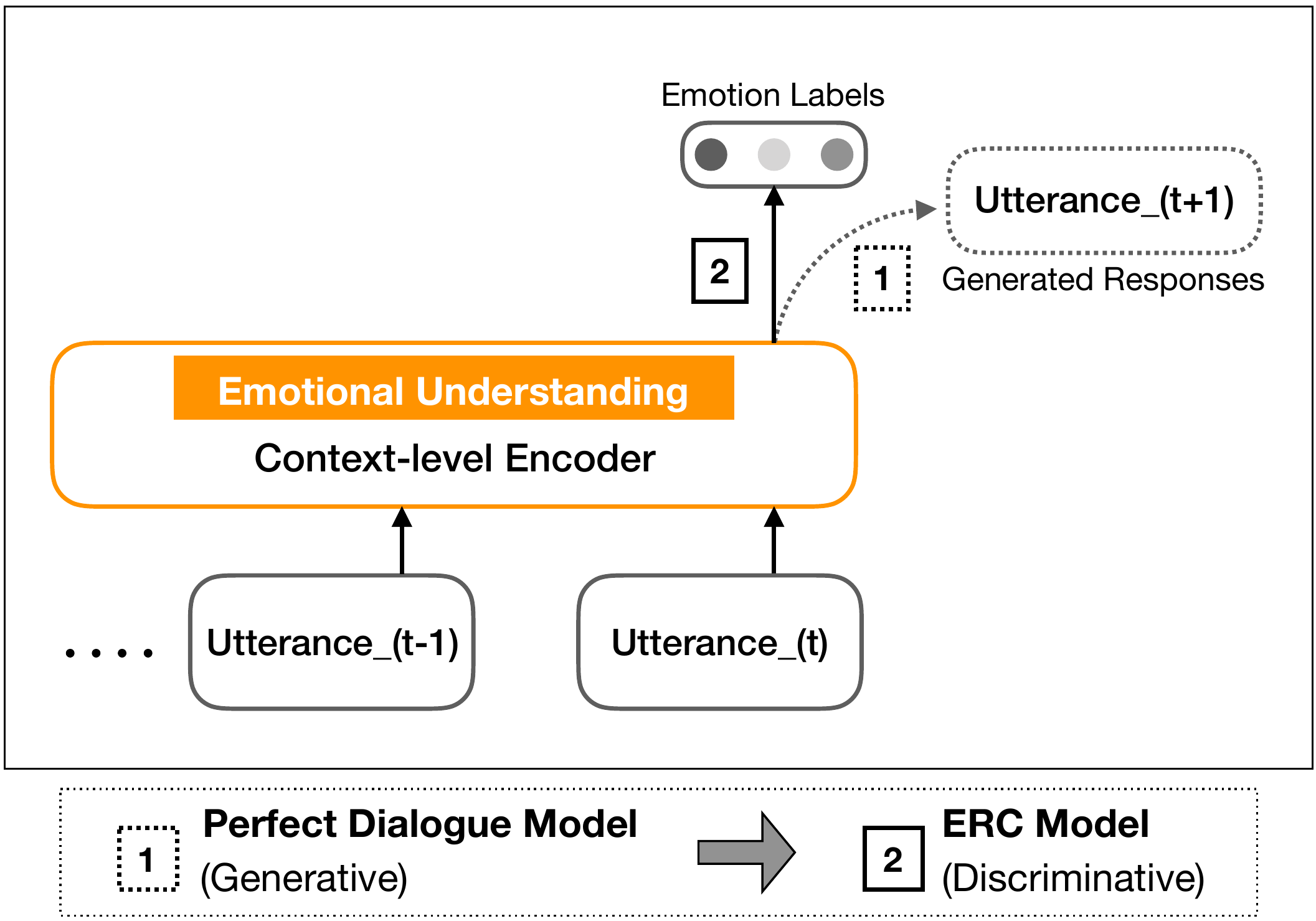}
	\caption{\footnotesize{The figure illustrates how a perfect dialogue generator requires emotional understanding from its context -- a transferrable knowledge into ERC.}}
	\label{fig:perfect_dialgoue_gen}
\end{figure}

To model such conversations, a generator would require the ability to 1) interpret latent emotions from its contextual turns and 2) model the complex dynamics governing them. In addition, it would also need to interpret other factors such as topic of the conversations, speaker personalities, intents, etc. Such a model would then be a \textit{perfect} dialogue generator. We illustrate this in \cref{fig:perfect_dialgoue_gen}, where the model generating utterance $utt_{t+1}$ would require to understand the emotions of the context arising from the utterances $utt_{t}, utt_{t-1},$ and so on. Thereby, we hypothesize that a trained dialogue generator would possess the ability to model implicit affective patterns across a conversation~\cite{shimizu2018pretraining}. Consequently, we propose a framework that uses TL to transfer this affective knowledge into our target discriminative task, i.e., ERC.

In our approach, we first pre-train a hierarchical generative dialogue model on the \textit{source task} of conversation modeling. Being an unsupervised (or self-supervised) task, conversation modeling typically benefits from a large amount of data in the form of multi-turn chats. Next, we \textit{adapt} our model to the \textit{target task} (ERC) by transferring the inter-sentence contextual parameters~\footnote{In this paper, \textit{context} refers to the inter-sentential context in conversations, i.e. the sequential information acquired from utterances of speakers in a conversation.} from the trained source model. For sentence encoding, we choose the BERT model~\cite{devlin2018bert}, which is pre-trained on masked language modeling and next sentence prediction objectives.

Although we acknowledge that training a \textit{perfect} dialogue generator is presently challenging, we demonstrate that benefits can be observed even with a popular baseline generator. In the bigger picture, our approach can enable the co-evolution of both generative and discriminative models for the tasks mentioned above. This is possible since improving an emotional classifier using a dialogue model can, in turn, be utilized to enhance dialogue models with emotional intelligence further, leading to an iterative cycle of improvements for both the applications.

Overall, our contributions are summarized as follows:

\begin{itemize}
	\item We propose \textit{TL-ERC}, which pre-trains a hierarchical generative dialogue model on multi-turn conversations (\textit{source}) and subsequently transfers affective knowledge to the \textit{target} task of ERC. Despite the active role of TL in providing state-of-the-art token and sentence encoders, its use in leveraging multi-turn contextual knowledge --- across utterances --- has been mostly unexplored. Our work stands as one of the first in this direction.
    \item Through our experiments, we observe the promising effects of using these pre-trained weights. Our models, initialized with the acquired knowledge, converge faster compared to randomly initialized counterparts and also demonstrate robust performance in limited training-data scenarios.
    \item We identify various challenges observed in using TL for ERC. These points raise essential research questions and provide a roadmap for future research in this topic.
\end{itemize}

In the remaining paper, \cref{sec:related_works} first discusses the works in the literature related to our task and our approach. Next, \cref{sec:model_design} provides information on the TL setup along with details on the design of the framework. Experimental details are mentioned in \cref{sec:experimental_setup}; results and extensive analyses are provided in \cref{sec:results}.~\cref{sec:challenges} provides some challenges observed in the proposed framework, casting light for future research efforts. Finally, \cref{sec:conclusion} concludes the paper.

\section{Related Works} \label{sec:related_works}

We proceed to discuss the use of transfer learning by the available literature in NLP. First, we enlist some of the famous works that have benefited from TL, and then we focus on works that attempt to frame TL in the context-level hierarchy. Next, we look at recent works on emotion/sentiment analysis, including works that have employed TL. Finally, we attempt to position our contribution amidst the latest developments in ERC.

\subsection{Transfer Learning in NLP}

Transfer learning has played a critical role in the success of modern-day NLP systems. As a matter of fact, the key milestones in the recent history of NLP are provided by works using TL. NLP has particularly benefited by inductive TL, where unlabelled data is utilized to leverage knowledge for labeled downstream tasks. Early works, such as~\citet{DBLP:journals/jmlr/AndoZ05}, introduced this concept, which was heavily adopted by the community and has shown tremendous success ever since~\cite{DBLP:conf/naacl/RuderPSW19}.

Modern breakthroughs, such as neural word embeddings, followed similar modeling by utilizing unlabeled textual data to learn the embeddings~\cite{mikolov2013distributed}. Of late, there has been a significant interest in using language models (LMs) to learn contextual embeddings~\cite{mccann2017learned,peters2018deep}. TL through LM pre-training has also provided state-of-the-art text classifiers with high quality sentence encoders~\cite{dai2015semi, devlin2018bert,DBLP:journals/corr/abs-1906-08237}. Consequently, several works have explored improving this framework by either modifying the LM pre-training approach or weight adaptation in the downstream tasks~\cite{howard2018universal,DBLP:journals/corr/abs-1907-11692}.

\paragraph{Context-level Transfer Learning}

Availability for works that explore TL for inter-sentence or sequential learning is limited. Some of these include sentence-level sequence tagging tasks~\cite{chen2019transfer} or inter-sentence supervised tasks such as query matching in conversations~\cite{qiu2018transfer}, next sentence prediction~\cite{devlin2018bert}, etc. Recent works that address the topic of pre-training sentence representations or multi-turn conversations follow either a retrieval-based or a generative strategy. For the former, strategies include contrastive sentence selection (ToD-BERT~\cite{DBLP:journals/corr/abs-2004-06871}, ConveRT~\cite{DBLP:journals/corr/abs-1911-03688}), sentence ordering (ALBERT~\cite{DBLP:conf/iclr/LanCGGSS20}), and semantical sentence matching (Sentence-BERT~\cite{DBLP:conf/emnlp/ReimersG19}) objectives. Whereas, generative models attempt to learn a probabilistic model for the conversations directly. DialoGPT~\cite{DBLP:journals/corr/abs-1911-00536} is a recently proposed model that proposes a generative model based on the GPT architecture~\cite{radfordlanguage}. Our pre-training model is similar in spirit to DialoGPT. However, we do not flatten the conversation and instead opt for a hierarchical conversation model. This also suits our downstream task of conversational emotion recognition. Additionally, we analyze the joint pre-training of full conversations in a self-supervised setting and attempt to observe its efficacy in transferring affective knowledge.

\subsection{Affect Analysis}

Affect, in particular emotions, are an integral part of human life and modulate our day-to-day behavior and activities~\cite{DBLP:journals/cim/CambriaPHL19}. The interest in understanding emotions is multi-disciplinary and covers a long history of research. The importance of modeling emotions has multiple benefits across applications such as e-learning~\cite{DBLP:journals/jnca/ImaniM19}, human-computer interaction~\cite{DBLP:conf/interspeech/LiscombeRH05}, user profiling~\cite{DBLP:series/lncs/SchiaffinoA09}, etc.

From a computational perspective, emotions are typically studied across various media formats, covering applications such as facial emotion recognition~\cite{DBLP:journals/corr/abs-1804-08348,DBLP:journals/ijon/WangPDZ18}, emotions in speech~\cite{DBLP:conf/webist/DrakopoulosPSP19,DBLP:journals/air/AnagnostopoulosIG15}, or multimodal emotion recognition~\cite{DBLP:series/lncs/MarechalMTPBAW19}. In text-based applications, machine learning has played a crucial role in mining emotions~\cite{DBLP:conf/naacl/AlmRS05}. Earlier approaches designed hand-crafted features that included emotional keyword spotting~\cite{DBLP:conf/sac/StrapparavaM08}, affect-based lexical resources (WordNet-Affect~\cite{DBLP:conf/lrec/StrapparavaV04}, SentiWordNet~\cite{DBLP:conf/lrec/Esuli006}), and distant supervision via hashtags~\cite{DBLP:conf/socialcom/0002CTS12}. In the present deep-learning era, 

In the present deep learning era, approaches have diverged from hand-crafted features and moved towards automated feature learning. Modern approaches consider advanced neural architectures, such as convolutional networks~\cite{choi-etal-2018-convolutional}, recurrent networks~\cite{DBLP:journals/corr/ChernykhSP17}, and attention mechanisms~\cite{DBLP:conf/icassp/MirsamadiBZ17} for emotion detection. Recent times have also seen approaches that address practical scenarios such as domain awareness~\cite{DBLP:journals/cim/Dragoni19}, and utilize alternate training strategies, such as adversarial approaches~\cite{DBLP:journals/cim/HanZS19}. Complementary to these issues, we address data scarcity issues in ERC and leverage transfer learning for the same. We discuss the related works aligned to these topics next.

\paragraph{Transfer Learning for Affect}

TL for affective analysis has gained momentum in recent years, with several works adopting TL-based approaches for their respective tasks. These works leverage diverse source tasks, such as, sentiment/emotion analysis in text~\cite{yu2018improving,DBLP:conf/semeval/Daval-FrerotBM18,bouchekif2019epita}, large-scale image classification in vision~\cite{ng2015deep}, sparse auto-encoding in speech~\cite{deng2013sparse}, etc. \citeauthor{DBLP:conf/emnlp/FelboMSRL17} utilize emojis present in online platforms to pre-train models and transfer knowledge for emotion recognition. Using layer-wise fine-tuning, they also transfer knowledge into related tasks of sarcasm and sentiment detection. A similar approach is taken by~\citeauthor{DBLP:conf/semeval/Daval-FrerotBM18}. Similar to these works, our approach also leverages TL for knowledge transfer. However, our task is in a sequential setting at the conversational level. To the best of our knowledge, our work is one of the first that explores TL in ERC and utilizes generative conversation modeling as a pre-training objective.
    
\paragraph{Emotion Recognition in Conversations} 

ERC is an emerging sub-field of affective computing and is developing into an active area of research. Current works try to model contextual relationships amongst utterances in a supervised fashion to model the implicit emotional dynamics. Strategies include modeling speaker-based dependencies using recurrent neural networks~\cite{DBLP:conf/semeval/Gonzalez-Garduno19,DBLP:conf/naacl/JiaoYKL19}, memory networks~\cite{DBLP:conf/naacl/HazarikaPZCMZ18,hazarika2018icon}, graph neural networks~\cite{DBLP:journals/corr/abs-1908-11540,DBLP:conf/ijcai/ZhangWSLZZ19}, quantum-inspired networks~\cite{DBLP:conf/ijcai/ZhangL0ZW19}, amongst others. Some of these works also explore challenges such as multi-speaker modeling~\cite{DBLP:conf/aaai/MajumderPHMGC19}, multimodal processing~\cite{hazarika2018icon}, and knowledge infusion~\cite{DBLP:journals/corr/abs-1909-10681}. BERT-based sentence encoding has also been heavily adopted by the latest works in this area~\cite{DBLP:conf/semeval/ChatterjeeNJA19}. Works like EmotionX-IDEA~\cite{DBLP:journals/corr/abs-1908-06264} and PT-Code~\cite{DBLP:journals/corr/abs-1910-08916}, developed concurrently to ours, follow a similar vein by transferring emotional knowledge from BERT pre-training. However, in these works, either the conversations are limited to utterance-reply pairs or follow a contrastive utterance retrieval objective. Our work, in contrast, pre-trains a whole conversation jointly using a hierarchical generative model. Overall, we find that there is a dearth of works that consider scarcity issues for annotated data and leverage TL. Our work strives to fill this gap by providing a systematic study for TL in ERC.

\section{Methodology} \label{sec:model_design}

\begin{figure*}[t]
	\centering
	\includegraphics[width=.8\linewidth]{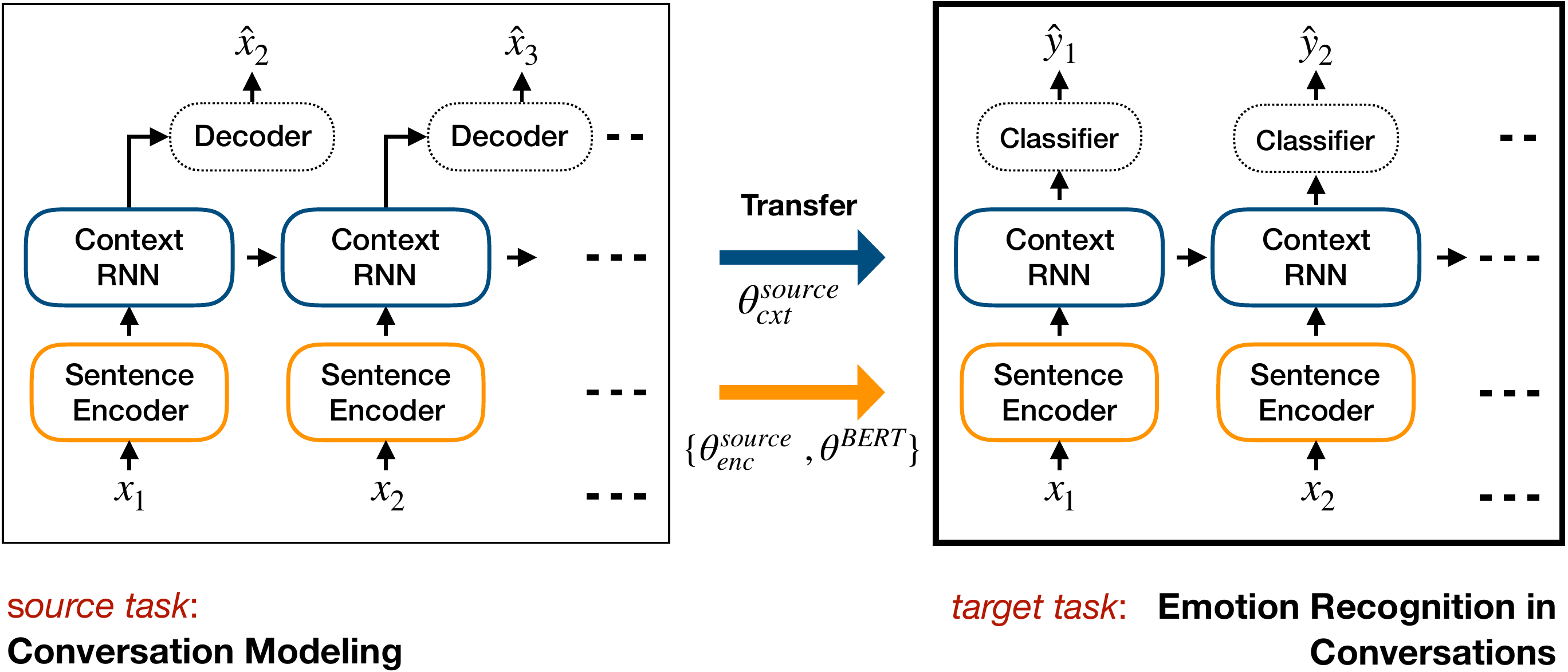}
	\caption{\footnotesize{Proposed framework for ERC using TL parameters. The model on the left is a conversational response generator which is used as a pre-trained model. The parameters are transferred to the target model as shown on the right side.}}
	\label{fig:framework}
\end{figure*}

Our proposed framework, TL-ERC, is summarized in \cref{fig:framework}. First, we define the source generative model trained as a dialogue generator, followed by a description of the target model, which performs hierarchical context encoding --- for the task of ERC --- using BERT-based sentence encoders and learned context weights from the source model.

\subsection{Source: Generative Conversation Modeling} To perform the generative task of conversation modeling, we use the \textit{Hierarchical Recurrent Encoder-Decoder} (HRED) architecture~\cite{serban2016building}. HRED is a classic framework for seq2seq conversational response generation that models conversations in a hierarchical fashion using three sequential components: \textit{encoder} recurrent neural networks (RNNs) for sentence encoding, \textit{context} RNNs for modeling the conversational context across sentences, and \textit{decoder} RNNs for generating the response sentence. 

For a given conversation context with sentences $x_1$, ... $x_{t}$, HRED generates the response $x_{t+1}$ as follows: 

	\begin{enumerate}[leftmargin=*]
		\item \textbf{Sentence Encoder:} It encodes each sentence in the context using an \textit{encoder RNN}, such that,
			\begin{align*}
				\textbf{h}_{t}^{enc} = f_{\theta}^{enc}(\textbf{x}_{t}, \textbf{h}_{t-1}^{enc})
			\end{align*}
		
		\item \textbf{Context Encoder:} The sentence representations are then fed into a \textit{context RNN} that models the conversational context until time step $t$ as 
			\begin{align*}
				\textbf{h}_t^{cxt} = f_{\theta}^{cxt}(\textbf{h}_{t}^{enc}, \textbf{h}_{t-1}^{cxt} )
			\end{align*}
		
		\item \textbf{Sentence Decoder:} Finally, an auto-regressive \textit{decoder RNN} generates sentence $x_{t+1}$ conditioned on $\textbf{h}_t^{cxt}$, i.e., 
			\begin{align*}
				p_{\theta}(x_{t+1}|x_{\leq t}) &= f_{\theta}^{dec}(x \ | \ \textbf{h}_t^{cxt}) \\ 
				&= \prod_{i}f_{\theta}^{dec}(x_{t+1,i} \ | \ \textbf{h}_t^{cxt}, x_{t+1,<i})
			\end{align*}
			
	\end{enumerate}

With the $i^{th}$ conversation being a sequence of utterances $C_i = [x_{i,1}, ..., x_{i,n_i}]$, HRED trains all the conversations in the dataset together by using the maximum likelihood estimation objective $\argmax_{\theta} = \sum_i{\log p_{\theta}(C_i)}$.

The HRED model provides the possibility to introduce multiple complexities in the form of multi-layer RNNs and other novel encoding strategies. In this work, we choose to experiment with the original version of the architecture with single-layer components so that we can analyze the hypothesis without unwanted contribution from the added complexities. In our source model, $f_{\theta}^{enc}$ can be any RNN function, which we model using the bi-directional \textit{Gated Recurrent Unit} (GRU) variant~\cite{cho2014learning} to encode each sentence. We call the parameters associated with this GRU function as $\theta_{enc}^{source}$. For both the \textit{context RNN} ($f_{\theta}^{cxt}$) and \textit{decoder RNN}, we use uni-directional GRUs --- with parameters $\theta_{cxt}^{source}$ and $\theta_{dec}^{source}$, respectively --- and complement the decoder with beam-decoding for generation~\footnote{Model implementations are adapted from \protect\url{https://github.com/ctr4si/}}.

\subsection{Target: Emotion Recognition in Conversations}

The input for this task is also a conversation $C$ with constituent utterances $[x_1, ..., x_n]$. Each $x_i$ is associated with an emotion label $y_i \in \mathbb{Y}$. We adopt a setup similar to the three components described for the source task, as in ~\citet{poria2017context}. However, the $decoder$ in this setup is replaced by a discriminative mapping to the label space instead of a generative network. Below, we describe the different initialization parameters that we consider for the first two stages of the network:

\subsubsection{Sentence Encoding} 
To encode each utterance in the conversation, we consider the state-of-the-art universal sentence encoder BERT~\cite{devlin2018bert}, with its parameters represented as $\theta^{BERT}$. We choose BERT over the HRED sentence encoder ($\theta_{enc}^{source}$) as it provides better performance (see~\cref{tab:glove_vs_bert}). Also, BERT includes the task of next sentence prediction as one of its training objectives which aligns with the inter-sentence level of abstraction that we consider in this work. 

We choose the \textit{BERT-base uncased} pre-trained model as our sentence encoder~\footnote{\url{https://github.com/huggingface/pytorch-pretrained-BERT}}. Though this model contains $12$ transformer layers, to limit the total number of parameters in our model, we restrict to the first 4 transformer layers. To get a sentential representation, we use the hidden vectors of the first token [CLS] across the considered transformer layers (see~\citet{devlin2018bert}) and mean-pool them to get the final sentence representation.

\subsubsection{Context Encoding} 
We use a similar context encoder RNN as the source HRED model with the option to transfer the learned parameters $\theta_{cxt}^{source}$. For input sentence representation $\textbf{h}_{t}^{enc}$ provided by the \textit{encoder RNN}, the \textit{context RNN} transforms it as follows:
	\begin{gather*}
	\textbf{z}_t = \sigma(V^z\textbf{h}_{t}^{enc} + W^z\textbf{h}_{t-1}^{cxt} + \textbf{b}^z) \\ \displaybreak[0]
	\textbf{r}_t = \sigma(V^r\textbf{h}_{t}^{enc} + W^r\textbf{h}_{t-1}^{cxt} + \textbf{b}^r) \\ \displaybreak[0]
	\textbf{v}_t = tanh(V^h\textbf{h}_{t}^{enc} + W^h(\textbf{h}_{t-1}^{cxt} \otimes \textbf{r}_{t}) + \textbf{b}^h) \\ \displaybreak[0]
	\textbf{h}_{t}^{cxt} = (1-\textbf{z}_{t})\otimes \textbf{v}_{t} + \textbf{z}_{t} \otimes \textbf{h}_{t-1}^{cxt} \\
	\textbf{h}_{t}^{cxt} = tanh(W^p\textbf{h}_{t}^{cxt} + \textbf{b}^p)
	\end{gather*}
	Here, $\{V^{z,r,h}, W^{z,r,h}, \textbf{b}^{z,r,h}$ are parameters for the GRU function and $\{W^p, \textbf{b}^p\}$ are additional parameters of a dense layer. For our setup, adhering to size considerations, we consider our transfer parameters to be $\theta_{cxt}^{source} = \{W^{z,r,h,p}, \textbf{b}^{z,r,h,p}\}$.

\subsubsection{Classification} 

For each turn in the conversation, the output from the context RNN is projected to the label-space, which provides the predicted emotion for the associated utterance. Similar to HRED, we train for all the utterances in the conversation together using the standard \textit{Cross Entropy} loss. For regression targets, we utilize the \textit{Mean Square Error (MSE)} loss, instead.

\section{Experimental Setup} \label{sec:experimental_setup}

In this section, we define the experimental setup followed in this work.
First, we detail the datasets that we utilize and mention their properties. Further, we provide information on the metrics used for evaluation, the training setup, and the model variants considered to test our hypothesis.

\subsection{Datasets}

\subsubsection{Source Task} 

For pre-training with the source task of conversation modeling, we consider two large-scale benchmark datasets: 
	
	\begin{itemize}
		\item  \textit{Cornell Movie Dialog Corpus}~\cite{danescu2011chameleons} is a popular collection of fictional conversations extracted from movie scripts. In this dataset, conversations are sampled from a diverse set of $617$ movies leading to over $83$k dialogues. 
        \item  \textit{Ubuntu Dialog Corpus}~\cite{lowe2015ubuntu} is a larger corpus with around 1 million dialogues, which, like the Cornell corpus, comprises of unstructured multi-turn dialogues based on Ubuntu chat logs (Internet Relay Chat).
	\end{itemize}
	
	Both datasets contain dyadic, i.e. two-party conversations. For brevity, throughout the paper, we mention these datasets as Cornell and Ubuntu, respectively. The data splits for training are created as per ~\citet{park2018hierarchical}.

\subsubsection{Target Task} 

For the target task of ERC, we experiment with three datasets popular in this area of research:
	
	\begin{itemize}
		\item Primarily, we consider the textual modality of a small-sized multimodal dataset \textit{IEMOCAP}~\cite{busso2008iemocap} consisting of dyadic conversations between 10 speakers. Each pair is assigned one of many diverse conversational scenarios, with a total of five sessions across the dataset. Each conversational video is segmented into utterances and annotated with the following emotion labels: \textit{anger, happiness, sadness, neutral, excitement,} and \textit{frustration}. Split creating scheme is based on~\citet{hazarika2018icon}.
		\item We also analyze results on a moderately-sized emotional dialogue dataset \textit{DailyDialog}~\cite{li2017dailydialog} with labeled emotions: \textit{anger, happiness, sadness, surprise, fear disgust} and \textit{no\_emotion}. Unlike spoken utterances in IEMOCAP, the conversations are chat-based based on daily life topics. For creating the splits, we follow the original split details provided by~\citet{li2017dailydialog}.
		\item Finally, we choose a regression-based dataset \textit{SEMAINE}, which is a video-based corpus of human-agent emotional interactions. We use the split configuration detailed in AVEC 2012's \textit{fully continuous sub-challenge}~\cite{schuller2012avec} for the prediction of affective dimensions: \textit{valence}, \textit{arousal}, \textit{power}, and \textit{expectancy}. Annotation configuration is based on~\citet{hazarika2018icon}.
	\end{itemize}
	
	\cref{tab:dataset_stats} provides the sizes along with split distributions for the above-mentioned datasets. For both IEMOCAP and SEMAINE, we generate the validation sets by random-sampling of 20\% dialogue videos from the training sets. The class distribution for the categorical emotions in IEMOCAP and DailyDialog are presented in \cref{tab:class_dist}. From the table, IEMOCAP is observed a fairly balanced dataset whereas DailyDialog is highly skewed towards sentences with no emotion. As such, we decide upon different metrics for each dataset as discussed next.

\begin{table}
\centering
\makebox[0pt][c]{\parbox{\textwidth}{%
    \begin{minipage}[t]{0.48\hsize}\centering
        
        \resizebox{0.95\linewidth}{!}{
    	\begin{tabular}{|llc|ccc|}
    		\hline
    	    &\multicolumn{2}{c|}{\multirow{2}{*}{Dataset}} & \multicolumn{3}{c|}{Dataset splits} \\
    		& && \small{train} & \small{validation} & \small{test} \\
    		\hline
    		\hline
    		\multirow{4}{*}{\rotatebox{90}{\small{Source}}}&\multirow{2}{*}{Cornell} &\small{\#D}& 66,477 & 8,310 & 8,310 \\
    		&&\small{\#U}& 244,030 & 30,436 & 30,247 \\
    		&\multirow{2}{*}{Ubuntu} &\small{\#D}& 898,142 & 18,920 & 19,560 \\
    		& &\small{\#U}&  6,893,060 & 135,747 & 139,775 \\
    		\hline
    		\multirow{4}{*}{\rotatebox{90}{\small{Target }}}&\multirow{2}{*}{IEMOCAP} &\small{\#D}& \multicolumn{2}{c}{120} & 31 \\
    		& &\small{\#U}& \multicolumn{2}{c}{5810} & 1,623 \\
    		\multirow{4}{*}{\rotatebox{90}{\small{Target }}}&\multirow{2}{*}{SEMAINE} &\small{\#D}& \multicolumn{2}{c}{58} & 22 \\
    		& &\small{\#U}& \multicolumn{2}{c}{4386} & 1,430 \\
    		&\multirow{2}{*}{Dailydialog} &\small{\#D}& 11,118 & 1,000 & 1,000 \\
    		&&\small{\#U}& 87,170 & 7,740 & 8,069 \\
    		\hline
    	\end{tabular}}
    	\caption{\footnotesize{Table illustrates the sizes of the datasets used in this work. \#D represents the number of dialogues whereas \#U represents the total number of constituting utterances.}}
    	\label{tab:dataset_stats}

    \end{minipage}
    \hfill
    \begin{minipage}[t]{0.48\hsize}\centering
        
        \resizebox{0.95\linewidth}{!}{

    	\begin{tabular}{|l|c:c|c:c:c|}
    		\hline
    	    &\multicolumn{2}{c|}{\textbf{Iemocap}} & \multicolumn{3}{c|}{\textbf{Dailydialog}}\\
    		&\small{train/val} & \small{test} & \small{train} & \small{val} & \small{test} \\
    		\hline
    		\hline
    		hap & 504 & 144 & 11182  & 684 & 1019\\
    		sad & 839 & 245 & 969 & 79  & 102\\
    		neu & 1324 & 384 & 72143 & 7108 & 6321\\
    		ang & 933 & 170 & 827 & 77 & 118 \\
    		exc & 742 & 299 & - & - & - \\
    		frus & 1468 & 381 & - & - & - \\
    		surp & - & -  & 1600  & 107 & 116 \\
    		fear & - & - & 146 &  11 & 17 \\
    		disg & - & - & 303  & 3 & 47 \\
    		\hline
    	\end{tabular}
    	}
    	\caption{\footnotesize{Category-wise distribution of utterances. \textit{hap}: happiness; \textit{neu}: neutral or no emotion; \textit{ang}: anger; \textit{exc}: excitement; \textit{frus}: frustration; \textit{surp}: surprise; \textit{disg}: disgust.}}
    	\label{tab:class_dist}
    	
    \end{minipage}
}}
\end{table}


\subsubsection{Metrics} We choose the pre-training weights from the source task based on the best validation perplexity score~\cite{park2018hierarchical}. For ERC, we use \textit{weighted-F-score} metric for the classification tasks on IEMOCAP and DailyDialog. For DailyDialog, we remove \textit{no\_emotion} class from the F-score calculations due to its high majority ($82.6\%$/$81.3\%$ occupancy in training/testing set) which hinders evaluation of other classes\footnote{Evaluation strategy adapted from Semeval 2019 ERC task: \protect\url{www.humanizing-ai.com/emocontext.html}}. For the regression task on SEMAINE, we take the Pearson correlation coefficient ($r$) as its metric.

We also provide the average \textit{best epoch} (BE) on which the least validation losses --- across the multiple runs --- are observed, and the testing evaluations are performed. A lower BE represents the model's ability to reach optimum performance in lesser training epochs.

\subsection{Model Size}
We consider two versions of the source generative model: \textbf{HRED-small} and \textbf{HRED-large} with $256$ and $1000$-dimensional hidden state sizes, respectively. While testing the performance of both the models on the IEMOCAP dataset, we find the context weights from \textit{HRED-small} (Cornell dataset) to provide better performance on average ($58.5\%$ F-score ) over \textit{HRED-large} ($55.3\%$ F-score). Following this observation, and also to avoid over-fitting on the small target datasets due to increased parameters, we choose the \textit{HRED-small} model as the source task model for our TL procedure.

\subsection{Model Variants and Baselines} \label{sec:baselines}

\begingroup
\renewcommand{\arraystretch}{1.3} 
\begin{table}[h]
	\centering
	\resizebox{0.9\linewidth}{!}{

	\begin{tabular}{|c|c:c|c|}
		\hline
	    \multirow{2}{*}{Variant}&\multicolumn{2}{c|}{\small{\textbf{Initial Weight}}} & \multirow{2}{*}{\small{\textit{Model Description}}}  \\
		& \multicolumn{1}{c:}{sent$_{enc}$} & cxt$_{enc}$ & \\ \hline \hline
		\multirow{2}{*}{$(1)$} & \multirow{2}{*}{-} & \multirow{2}{*}{-} & \small{Sentence encoders -- \textit{randomly} initialized.} \\
		& & &\small{Context encoders -- \textit{randomly} initialized.}\\ 
		\hline
		
		\multirow{2}{*}{$(2)$} & \multirow{2}{*}{$\theta^{BERT}$} & \multirow{2}{*}{-} & \small{Sentence encoders -- BERT parameters.} \\
		& & & \small{Context encoders -- \textit{randomly} initialized.}\\ 
		\hline
		\multirow{4}{*}{$(3)$} & \multirow{4}{*}{$\theta^{BERT}$} & \multirow{4}{*}{$\theta^{ubuntu/cornell}_{cxt}$} & \small{\textbf{TL-ERC}}   \\
		& & & \small{Sentence encoders -- BERT parameters.}\\
		& & & \small{Context encoders -- initialized from generative models}\\
		& & & \small{pre-trained on Ubuntu/Cornell corpus.}\\
		\hline
		
	\end{tabular}
	}
	\caption{\footnotesize{Variants of the model used in the experiments. Variant (3) is the proposed TL-ERC model.}}
	\label{tab:model_variants}
\end{table}
\endgroup

The primary goal of this paper is to analyze the effect of TL at the conversation level for ERC. For this, we experiment on different variants of our model based on the parameter initialization procedure. We provide a summary of these variants in~\cref{tab:model_variants}. In the table, \textit{Variant 1} is the model with randomly initialized parameters. In \textit{Variant 2}, we replace the sentence encoders with the BERT model including its original pre-trained parameters. Finally, in \textit{Variant 3}, in addition to BERT sentence encoders, we also initialize the context-RNN parameters learned from the source task. Different results and analyses amongst these variants are provided in~\cref{sec:results}.

Next, to compare our model with the existing literature, we select some prior state-of-the-art models evaluated on the target datasets:

\begin{itemize}[leftmargin=*]
    \item CNN~\cite{DBLP:conf/emnlp/Kim14} extracts textual features based on Convolutional Neural Networks (CNN). This is a non-contextual model, which evaluates each utterance in a conversation independently.
    
    \item Memnet~\cite{DBLP:conf/nips/SukhbaatarSWF15} assigns dedicated memory for each historical utterance and performs multi-hop inference on them to get final representations for emotion classification.
    
    \item c-LSTM~\cite{poria2017context} is a popular model which is similar to our target model. It employs a bi-directional LSTM~\cite{DBLP:journals/neco/HochreiterS97} to capture inter-utterance dependencies.
    
    \item c-LSTM+Att~\cite{DBLP:conf/icdm/PoriaCHMZM17} enhances the c-LSTM model with inter-modality and inter-utterance attention mechanisms.
    
    \item CMN~\cite{DBLP:conf/naacl/HazarikaPZCMZ18}, the Conversational Memory Network, is an extension to the Memnet model which allots separate memories to both speakers in a dyadic conversational exchange.
    
    \item DialogueRNN~\cite{DBLP:conf/aaai/MajumderPHMGC19} is a strong state-of-the-art baseline which employs three stages of recurrent units comprising global, speaker-state, and emotional units. The global RNN models the conversational context, speaker-state RNN models the individual speaker-states, and emotion RNN models the final emotional representations used for classification. For a fair comparison with our model, we chose the basic version of DialogueRNN without bi-directional RNNs and inter-utterance attention mechanisms.
\end{itemize}

\noindent Results on these baselines are provided in~\cref{sec:baseline_results}.

\subsection{Training Criteria}

\paragraph{Hyper-parameter search} For each target dataset-model combination, we perform grid-search to select the appropriate hyper-parameters. In the search procedure, we keep the model architecture constant but vary learning rate (1e-3, 1e-4, and 1e-5), optimizer (Adam, RMSprop~\cite{Tieleman2012}), batch size (2-40 videos/batch), and dropout (\{0.0, 0.5\}. BERT-parameters contain dropout of 0.1 as in \citet{devlin2018bert}). For a particular dataset-model pair, the final hyper-parameter configuration is chosen based on the best performance on the respective validation set. In the case of negligent difference between the combinations, we use the Adam optimizer~\cite{kingma2014adam} as the default variant with $\beta = [0.9, 0.999]$ and learning rate $1e-4$.

\paragraph{Inference} 
We train our models on each target dataset for multiple runs (10:IEMOCAP, 5:DailyDialog, 5:SEMAINE). In each run,
the training proceeds with an \textit{early stopping} criterion of patience 10. During this training loop, the parameters with the least validation loss are finally chosen for the testing-set inference and evaluation.

\begingroup
\renewcommand{\arraystretch}{1.3} 
\begin{table*}[t]
	\centering
	\resizebox{\linewidth}{!}{

	\begin{tabular}{|c|cc|cc|cc|cc|cc|}
		\hline
	    \parbox[t]{2mm}{\multirow{3}{*}{\rotatebox[origin=c]{90}{Variant}}} & & & \multicolumn{8}{c|}{Dataset: \textbf{IEMOCAP}} \\
	    &\multicolumn{2}{c|}{\small{\textbf{Initial Weights}}}  & \multicolumn{2}{c|}{$10\%$} & \multicolumn{2}{c|}{$25\%$} & \multicolumn{2}{c|}{$50\%$} & \multicolumn{2}{c|}{$100\%$}\\
		& sent$_{enc}$ & cxt$_{enc}$ & \small{F-Score} & \small{BE} & \small{F-Score} & \small{BE} & \small{F-Score} & \small{BE} & \small{F-Score} & \small{BE} \\
		\hline
		\hline
		(1) & - & -  & 23.2 $_{\pm 0.4}$   & 48.4 & 41.6 $_{\pm 0.8}$ & 72.5  & 48.4 $_{\pm 0.3}$ & 75.1 & 53.8  $_{\pm 0.3}$ & 13.8  \\ \cline{1-3}
		(2) & \multirow{1}{*}{$\theta^{BERT}$} & - & 32.4 $_{\pm 1.1}$ & \textbf{11.0} & 41.9 $_{\pm 0.5}$ & \textbf{8.0} & 49.2 $_{\pm 1.0}$ & \textbf{6.3} & 55.1 $_{\pm 0.6}$ & \textbf{5.0} \\ 
		\cline{1-3}
		\multirow{2}{*}{(3)} & \multirow{2}{*}{$\theta^{BERT}$} & $\theta^{ubuntu}_{cxt}$  & 35.7 $_{\pm 1.1}$ & 14.2 &  45.9 $_{\pm 2.0}$ & 11.2  & \textbf{53.1} $_{\pm 0.7}$$^{\dagger}$  & 7.8  & \textbf{58.8 $_{\pm 0.5}$}$^{\dagger}$  & 5.4  \\
		 & & $\theta^{cornell}_{cxt}$  & \textbf{36.3 $_{\pm 1.1}$}$^{\dagger}$ & 17.0 & \textbf{46.0} $_{\pm 0.5}$$^{\dagger}$ & 11.2 & 50.9 $_{\pm 1.5}$ & 8.2 & 58.5 $_{\pm 0.8}$ & \textbf{5.0}\\
		\hline
	\end{tabular}
	}
	\caption{\footnotesize{IEMOCAP results. Metric: Weighted-Fscore averaged over 10 random runs. BE = Best Epoch. Results span across different amount of available training data. Validation and testing splits are fixed across configurations. $^{\dagger}$ represents significant difference with $p < 0.05$ over randomly initialized model as per two-tailed Wilcoxon rank sum hypothesis test~\cite{nachar2008mann}.}}
	\label{tab:iemocap_results}
\end{table*}
\endgroup

\begingroup
\renewcommand{\arraystretch}{1.3} 
\begin{table}[t]
	\centering
	\resizebox{0.7\linewidth}{!}{

	\begin{tabular}{|c|cc|cc|cc|}
		\hline
	    \parbox[t]{2mm}{\multirow{3}{*}{\rotatebox[origin=c]{90}{Variant}}} & & & \multicolumn{4}{c|}{Dataset: \textbf{DailyDialog}} \\
	    & \multicolumn{2}{c|}{\small{\textbf{Initial Weights}}} & \multicolumn{2}{c|}{$10\%$} & \multicolumn{2}{c|}{$100\%$}\\
		& sent$_{enc}$ & cxt$_{enc}$ &  \small{F-score} &\small{BE} & \small{F-score} &\small{BE}  \\
		\hline
		\hline
		(1) & - & -  &  33.5  $_{\pm 2.2}$  & 12.3  &  45.3 $_{\pm 1.9}$ & 7.9  \\ \cline{1-3}
		(2) & \multirow{1}{*}{$\theta^{BERT}$}  & - &  37.5  $_{\pm 1.8}$  &\textbf{ 2.6} &  47.4 $_{\pm 1.2}$ & \textbf{2.4}\\ \cline{1-3}
		\multirow{2}{*}{(3)} & \multirow{2}{*}{$\theta^{BERT}$} & $\theta^{ubuntu}_{cxt}$  & 37.7  $_{\pm 3.1}$  & 3.1 & 47.1 $_{\pm .76}$  & \textbf{2.4} \\
	    & & $\theta^{cornell}_{cxt}$  & \textbf{38.5} $_{\pm 1.5}$$^{\dagger}$  & 3.2 &  \textbf{48.0} $_{\pm 1.8}$$^{\dagger}$ & \textbf{2.4} \\
		\hline
	\end{tabular}
	}
	\caption{\footnotesize{DailyDialog results. Metric: Weighted-Fscore averaged over 5 random runs. BE = Best Epoch. $^{\dagger}$ represents significant difference with $p < 0.05$ over random initialized model as per two-tailed Wilcoxon rank sum hypothesis test~\cite{nachar2008mann}.}}
	\label{tab:dailydialog_results}
\end{table}
\endgroup

\begingroup
\renewcommand{\arraystretch}{1.3} 
\begin{table}[t]
	\centering
	\resizebox{0.8\linewidth}{!}{

	\begin{tabular}{|c|cc|cc|cc|cc|cc|}
		\hline
	    \parbox[t]{2mm}{\multirow{3}{*}{\rotatebox[origin=c]{90}{Variant}}} & \multicolumn{2}{c|}{\small{\textbf{Initial Weights}}} & \multicolumn{8}{c|}{Dataset: \textbf{SEMAINE}} \\
	    & & & \multicolumn{2}{c}{DV} &  \multicolumn{2}{c}{DA}  &   \multicolumn{2}{c}{DP}  &   \multicolumn{2}{c|}{DE}\\
		& sent$_{enc}$ & cxt$_{enc}$ &  \small{r} &\small{BE} &  \small{r} &\small{BE} &  \small{r} &\small{BE} &  \small{r} &\small{BE}\\
		\hline
		\hline
		(1) & - & -  & 0.14  & \textbf{4} & 0.27 & 6.2 & 0.18 & 12.8 & -0.03 & 287.4\\ \cline{1-3}
		(2) & \multirow{1}{*}{$\theta^{BERT}$}  & - & 0.64  & 13.8 & 0.36  & 7.8 &  0.33  & 4.8 &  -0.03 &  23   \\ \cline{1-3}
	    \multirow{2}{*}{(3)} & \multirow{2}{*}{$\theta^{BERT}$}  & $\theta^{ubuntu}_{cxt}$  & \textbf{0.66}  & 10.2 & 0.41 & \textbf{6} & 0.34  & 3.8 &  -0.03 &   23   \\
	    & & $\theta^{cornell}_{cxt}$  & 0.65  & 10.2 & \textbf{0.42} & 8.8 & \textbf{0.35} & \textbf{3.4} & -0.029  &\textbf{22.7}     \\
		\hline
	\end{tabular}
	}
	\caption{\footnotesize{SEMAINE results. Metric (r): Pearson correlation coefficients averaged over 5 random runs. DV = Valence, DA = Activation/Arousal, DP = Power, DE =Anticipation/Expectation.}}
	\label{tab:avec_results}
\end{table}
\endgroup

\section{Results and Analyses} \label{sec:results}

\cref{tab:iemocap_results} and \ref{tab:dailydialog_results} provide the performance results of ERC on classification datasets IEMOCAP and DailyDialog, respectively. In both the tables, we observe clear and statistically significant improvements of the models that use pre-trained weights over the randomly initialized variant. We see further improvements when context-modeling parameters from the source task ($\theta_{cxt}^{source}$) are transferred, indicating the benefit of using TL in this context-level hierarchy.

Similar trends are observed in the regression task based on the SEMAINE corpus (see~\cref{tab:avec_results}). For \textit{valence}, \textit{arousal}, and \textit{power} dimensions, the improvement is significant. For \textit{expectation}, the performance is marginally better but at a much lesser BE, indicating faster generalization.

In the following sections, we take a closer look at various aspects of our approach that include checking robustness towards limited-data scenarios, generalization time, and questioning design choices. We also provide additional analyses that probe the existence of data-split bias, domain influence, and effect of fine-tuning strategies.

\subsection{Target Data Size}

Present approaches in ERC primarily adopt supervised learning strategies that demand a high amount of annotated data. However, the publicly available datasets in this field belong to the small-to-medium range in the spectrum of dataset sizes. For example, other applications of NLP have datasets of much larger sizes -- over $130k$ instances in SQuAD for Question Answering~\cite{DBLP:conf/acl/RajpurkarJL18}, over $393k$ instances in MNLI for language inference~\cite{N18-1101}, and so on. This constraint inhibits the true potential of systems trained on these datasets. As a result, approaches that provide higher performance in a limited training-data scenario tend to be highly desirable, particularly for ERC.

We design experiments to check the robustness of our models against such limited settings. To limit the amount of available training data, we create random subsets of the training dialogues while maintaining the original label-distribution. In both \cref{tab:iemocap_results} and \ref{tab:dailydialog_results}, we observe that the pre-trained models are significantly more robust against limited training resources compared to models trained from scratch.

\paragraph{Effect of bias in random splits}

We investigate the possibility of bias in the random splits, which aid in supporting our hypothesis. To eliminate this possibility, we further check if the improvement in our TL-based approach --- for the limited-data scenarios --- are triggered by such data-split bias. In other words, we pose the following question, \textit{if another training split is sampled from the original dataset, would our model provide similar improvements?} We provide evidence that this is indeed true.

\cref{tab:split_comparisons} presents the results where for $10\%$ and $50\%$ training-data setup, we sample $4$ independent splits from the IEMOCAP dataset. As seen from the table, different splits provide different results, which is expected owing to the variances in the samples and their corresponding labels. However, the relative performance within each split follows similar trends of improvement for TL-based models. This observation nullifies the potential existence of bias in the reported results.

\subsection{Target Task's Training Time}

\begin{figure}[h]
    \centering
	\includegraphics[width=0.5\linewidth]{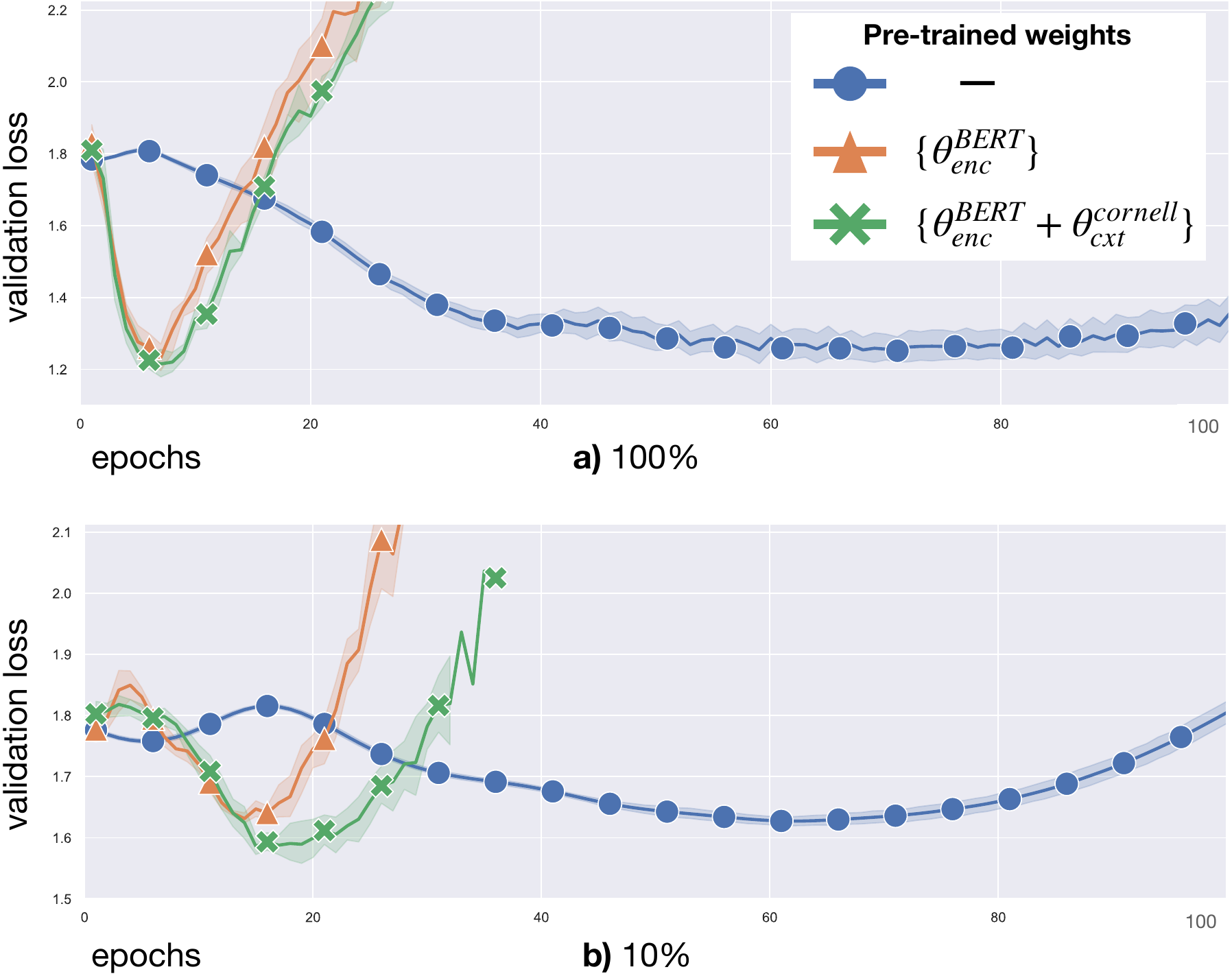}
	\caption{\footnotesize{Validation loss across epochs in training for different weight-initialization settings on the IEMOCAP dataset. Part a) represents results when trained on $100\%$ training data b) $10\%$ training data split. For fair comparison, optimizer learning rates are fixed at 1e-4.}}
	\label{fig:val_loss}
\end{figure}

In all the configurations in \cref{tab:iemocap_results} and \ref{tab:dailydialog_results}, we observe that the presence of  weight initialization leads to faster convergence in terms of the best validation loss. \cref{fig:val_loss} demonstrates the trace of the validation loss on training data configurations of the IEMOCAP dataset. As observed, the pre-trained models achieve their best epoch in a significantly shorter time which indicates that the transferred weights are helping the model better guide to its optimal performance. 

\subsection{Encoder Initialization}

\cref{tab:glove_vs_bert} provides a comparative study between the performance of models initialized with HRED-based sentence encoders ($\theta_{enc}^{source}$) versus the BERT encoders ($\theta^{BERT}$) that we use in our final networks. Results demonstrate that BERT provides better representations, which leads to better performance. Moreover, the positive effects of the context parameters are observed when coupled with the BERT encoders. This behavior indicates that the performance boosts provided by the context-encoders is contingent on the quality of sentence encoders. Observing this empirical evidence, we choose BERT-based sentence encoders in our final network.

\begingroup
\renewcommand{\arraystretch}{1.3} 
\begin{table}[t]
    \centering
    \resizebox{\linewidth}{!}{
	\begin{tabular}{|c|cc|c:c:c:c|c:c:c:c|}
		\hline
	    \parbox[t]{2mm}{\multirow{3}{*}{\rotatebox[origin=c]{90}{Variant}}}&&&\multicolumn{8}{c|}{Dataset: \textbf{IEMOCAP}} \\
	    &\multicolumn{2}{c|}{\small{\textbf{Initial Weight}}}  & \multicolumn{4}{c|}{$10\%$}& \multicolumn{4}{c|}{$50\%$}\\
		&\multirow{1}{*}{sent$_{enc}$} & \multirow{1}{*}{cxt$_{enc}$} & split$_{1}^{*}$ & split$_{2}$ & split$_{3}$ & split$_{4}$ & split$_{1}^{*}$ & split$_{2}$ & split$_{3}$ & split$_{4}$ \\ \cline{3-10}
		\hline
		\hline
		(1) &- & -  & 23.2 $_{\pm 0.4}$ & 31.5 $_{\pm 0.6}$ & 25.0 $_{\pm 1.7}$ & 8.8 $_{\pm 1.1}$ & 48.4 $_{\pm 0.3}$ & 48.5 $_{\pm 1.3}$ & 49.1 $_{\pm 0.9}$ & 51.3 $_{\pm 0.5}$ \\ \cline{1-3}
		(2) &\multirow{1}{*}{$\theta^{BERT}$} & - &  32.4 $_{\pm 1.1}$  &  31.6 $_{\pm 1.2}$  & 30.5 $_{\pm 0.8}$ & 23.65 $_{\pm 1.3}$ & 49.2 $_{\pm 1.0}$ & 49.0 $_{\pm 0.7}$ & 48.8 $_{\pm 0.9}$ & 51.4 $_{\pm 0.6}$ \\ \cline{1-3}
		\multirow{2}{*}{(3)} & \multirow{2}{*}{$\theta^{BERT}$} & $\theta^{ubuntu}_{cxt}$ & 35.7 $_{\pm 1.1}$ & 32.0 $_{\pm 1.1}$ & \textbf{39.0} $_{\pm 0.2}$ & \textbf{24.90} $_{\pm 3.0}$ & \textbf{53.1} $_{\pm 0.7}$  & 53.2 $_{\pm 1.3}$  & 52.9 $_{\pm 1.9}$ & 54.2 $_{\pm 0.8}$ \\
		& & $\theta^{cornell}_{cxt}$ & \textbf{36.3} $_{\pm 1.1}$  & \textbf{34.2} $_{\pm 0.8}$ & 35.7 $_{\pm 0.5}$ & 24.70 $_{\pm 1.2}$ & 50.9 $_{\pm 1.5}$  & \textbf{54.3} $_{\pm 0.8}$ & \textbf{ 53.5} $_{\pm 0.6}$ & \textbf{55.4} $_{\pm 1.0}$ \\
		\hline
		\multicolumn{6}{l}{\footnotesize{$^{*}$ primary split}}\\
	\end{tabular}
	}
	
	\caption{\footnotesize{Table to investigate if split randomness incurs bias in results. Comparisons are held between two limited training data scenarios comprising $10\%$ and $50\%$ available training data. For both the cases, $4$ independent splits are sampled and compared against. Metric: Weighted-Fscore averaged over 10 random runs.}}

    \label{tab:split_comparisons}
\end{table}
\endgroup

\begin{table}[h]
	\centering
	\resizebox{0.5\linewidth}{!}{

	\begin{tabular}{|cc|c|c|}
		\hline
	    & & \multicolumn{2}{c|}{Dataset: \textbf{IEMOCAP}} \\
	    \multicolumn{2}{|c|}{\small{\textbf{Initial Weight}}} & $10\%$ & $100\%$\\
		sent$_{enc}$ & cxt$_{enc}$ &  \small{F-score} & \small{F-score}  \\
		\hline
		\hline
		- & -  &  23.2 $_{\pm 0.4}$   &  53.8  $_{\pm 0.3}$ \\ \hline
		\multirow{2}{*}{$\theta^{cornell}_{enc}$} & - & 26.3 $_{\pm 0.9}$  &  54.9 $_{\pm 0.3}$ \\ 
		 & $\theta^{cornell}_{cxt}$  &  27.5 $_{\pm 1.3}$ & 55.1 $_{\pm 0.9}$ \\
		\hline
		\multirow{2}{*}{$\theta^{ubuntu}_{enc}$} & - & 24.6 $_{\pm 0.9}$ & 53.2 $_{\pm 0.5}$\\ 
		 & $\theta^{ubuntu}_{cxt}$  &  23.3 $_{\pm 0.8}$ & 53.7 $_{\pm 0.9}$ \\
		\hline
		\multirow{3}{*}{$\theta^{BERT}$} & - &  32.4 $_{\pm 1.1}$  &   55.1 $_{\pm 0.6}$ \\
		 & $\theta^{ubuntu}_{cxt}$  & 35.7 $_{\pm 1.1}$ &  \textbf{58.8 $_{\pm 0.5}$}  \\
	     & $\theta^{cornell}_{cxt}$  & \textbf{36.3 $_{\pm 1.1}$} &  58.5 $_{\pm 0.8}$ \\ \hline
		
	\end{tabular}
	}
	\caption{\footnotesize{Table to analyze HRED encoder vs BERT. Metric: Weighted-Fscore averaged over 10 random runs. BE = Best Epoch (average). }}
	\label{tab:glove_vs_bert}
\end{table}

\subsection{Impact of Source Domain.}

\begin{figure}[h]
    \centering
	\includegraphics[width=0.6\linewidth]{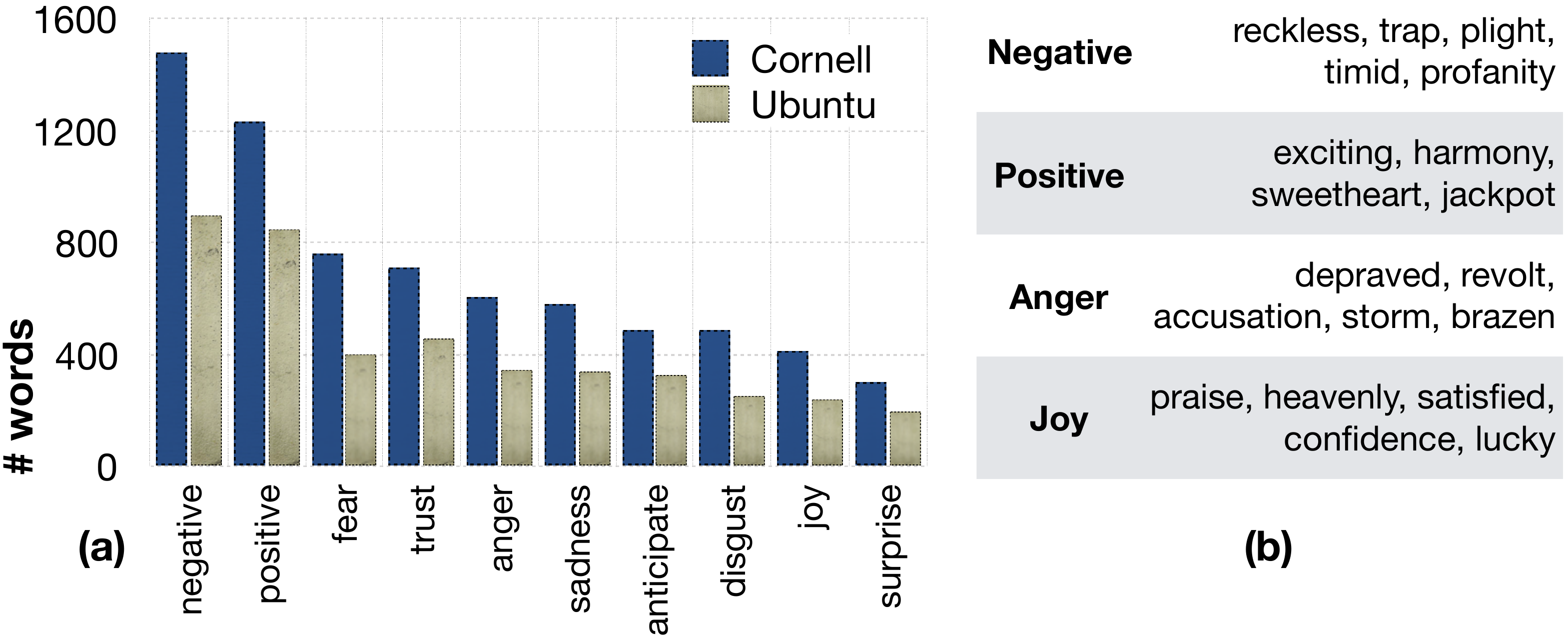}
	\caption{a) Frequency of emotive words from source datasets: Cornell and Ubuntu. b) randomly sampled words from Cornell associated to mentioned emotions.}
	\label{fig:emotion_color}
\end{figure}

We investigate if the choice of source datasets incur any significant change in the results. First, we define an emotional profile for the source datasets and observe whether any correlation is found between their emotive content versus the performance boost achieved by pre-training on them.

To set up an emotional profile, we look at the respective vocabularies of both corpora. For each token, we check its association with any emotion by using the emotion-lexicon provided by~\citet{mohammad2013crowdsourcing}. The NRC Emotion Lexicon contains 6423 words belonging to emotion categories: \textit{fear, trust, anger, sadness, anticipation, joy, surprise}, and \textit{disgust}. It also assigns two broad categories: \textit{positive} and \textit{negative} to describe the type of connotation evoked by the words. We enumerate the frequency of each emotion category amongst the tokens of the source dataset's vocabulary. To compose the vocabulary of both the source datasets, we set a minimum frequency threshold of 5, which provides $13518$ and $18473$ unique tokens for Cornell and Ubuntu, respectively. Each of the unique tokens is then lemmatized~\footnote{\protect\url{https://www.nltk.org/_modules/nltk/stem/wordnet.html}} and cross-referenced with the lexicon, which provides $3099$ (Cornell) and $2003$ (Ubuntu) tokens with associated emotions.

\cref{fig:emotion_color} presents the emotional profiles,  which indicate that the Cornell dataset has a higher number of emotive tokens in its vocabulary. However, the results illustrated in \cref{tab:iemocap_results}, \ref{tab:dailydialog_results}, and \ref{tab:avec_results} \textit{do not} present any significant difference between the two sources. A possible reason for this behavior attributes to the fact that such emotional profile relies on surface emotions derived from the vocabularies. However, as per our hypothesis, response generation includes emotional understanding as a latent process. This reasoning leads us to believe that surface emotions need not necessarily correlate to performance increments. Rather, the quality of generation would include such properties intrinsically.



\begin{table}
\centering
\makebox[0pt][c]{\parbox{\textwidth}{%
    \begin{minipage}[t]{0.38\hsize}\centering
        
        \resizebox{0.95\linewidth}{!}{

    	\begin{tabular}{|l|c|c|}
    		\hline
    	    \multicolumn{1}{|c|}{\small{Adapt Strategy}} &\multicolumn{1}{c|}{\textbf{Iemocap}} & \multicolumn{1}{c|}{\textbf{DD}}\\
    		\small{\textbf{Fixed weights}} &  \small{F-Score} & \small{F-Score}   \\
    		\hline
    		\hline
    		- & \textbf{58.5} &  \textbf{48.0} \\
    		$\theta^{BERT} $ & 17.0 & 32.1 \\
    	    $\theta^{BERT} + \theta^{cornell}_{cxt}$  & 9.3 &  4.5 \\
    		\hline
    	\end{tabular}
    	}
    	\caption{\footnotesize{Average performance on ERC with pre-trained weights: $\{\theta^{BERT}+\theta^{cornell}_{cxt}\}$. Note: DD here means DailyDialog.}}
    	\label{tab:freezing_results}

    \end{minipage}
    \hfill
    \begin{minipage}[t]{0.58\hsize}\centering
        
        \resizebox{0.95\linewidth}{!}{

    	\begin{tabular}{|l|c|c:c:c:c|}
    		\hline
    	    &\multicolumn{1}{c|}{\textbf{Iemocap}} & \multicolumn{4}{c|}{\textbf{SEMAINE}}\\
    	     &   & \small{DV} & \small{DA} & \small{DP} & \small{DE} \\
    		\small{\textbf{Models}} &  \small{F-Score} & \small{r} & \small{r} & \small{r} & \small{r} \\
    		\hline
    		\hline
    		CNN & 48.1 & -0.01 & 0.01 & -0.01 & 0.19 \\
    		Memnet & 55.1 & 0.16 & 0.24 & 0.23 & 0.05 \\
    		c-LSTM & 54.9 & 0.14 & 0.23 & 0.25 & -0.04 \\
    		c-LSTM + Att & 56.1 & 0.16 & 0.25 & 0.24 & 0.10 \\
    		CMN & 56.1 & 0.23 & 0.29 & 0.26 & -0.02 \\
    		DialogueRNN & \textbf{59.8} &  0.28 & 0.36 & 0.32 & \textbf{0.31} \\ 
    		\hline
    	    TL-ERC  & 58.8 & \textbf{0.66} & \textbf{0.42} & \textbf{0.35} & -0.02 \\
    		\hline
    	\end{tabular}
    	}
    	\caption{\footnotesize{Average performance of TL-ERC compared to previous state-of-the-art models.}}
    	\label{tab:previous_work}
    	
    \end{minipage}
}}
\end{table}

\subsection{Comparison with previous work.} \label{sec:baseline_results}

\cref{tab:previous_work} provides the results for various baselines detailed in~\cref{sec:baselines}. As seen, our proposed TL-ERC comfortably outperforms both non-contextual and contextual baselines. It achieves this without the aid of attention mechanisms that is used in c-LSTM + Att, multi-hop memory networks used in Memnet, and CMN. It also achieves competitive performance against DialogueRNN, which has three layers of inter-utterance recurrent layers, while TL-ERC has one. These trends indicate TL to be effective in our setup and provided promising directions for future research.

\section{Challenges} \label{sec:challenges}

In this section, we enlist the different challenges that we observed while experimenting with the proposed idea. These challenges provide roadmaps for further research on this topic to build better and robust systems.

\subsection{Adaptation Strategies}

We try two primary adaptation techniques used in inductive TL, \textit{freezed} or \textit{fine-tuned}. In the former setting, the borrowed weights are used for feature extraction, while in the latter,  we train the weights along with the other new parameters of the target task's model. Fine-tuning can also be performed using other techniques such as gradual unfreezing~\cite{peters2019tune}. In \cref{tab:freezing_results}, we experiment with freezing different amounts of transferred weights in our ERC model. We notice a degradation in performance with more frozen parameters. The datasets in ERC contain multi-class annotations with varying label distributions. With frozen parameters, our transferred model is unable to account for the label distribution and results in low recall for infrequent classes. We thus find the fine-tuning approach to be more effective in this setup.

However, fine-tuning all parameters also present higher susceptibility to over-fitting~\cite{howard2018universal}. We observe this trait in \cref{fig:val_loss}, where the validation loss shoots up at a faster rate than the random counterpart. Finding a fine-balance in this trade-off remains an open problem.

\subsection{Stability Issues}

In the results, we observe that the variability of the models across multiple runs (in terms of the standard error) is relatively higher for the proposed models as compared to randomly initialized weights. Though, on average, our models perform significantly better, there remains a scope for improvement to achieve more stable training.

\subsection{Variational Models} 

Many works utilize variational networks to model the uncertainties and variability in latent factors. For dialogue modeling, networks such as VHRED~\cite{SerbanSLCPCB17} incorporate such variational properties to model its latent processes. Emotional perception, in particular, has been argued to contain shades of multiple affective classes instead of a hard label assignment~\cite{DBLP:journals/taslp/MowerMN11}. We, thus, posit that variational dialogue models such as VHRED also hold the potential for improving affective knowledge transfer. 

We experiment on this concept by using VHRED as the source model. VHRED uses additional paramteres to model its prior latent state $\bm{z}_t$, which is then concatenated with $\bm{h}_t^{cxt}$ as follows:

\begin{gather*}
\textbf{h}_{t}^{enc} = f_{\theta}^{enc}(\textbf{x}_t) \\ \displaybreak[0]
\textbf{h}_{t}^{cxt} = f_{\theta}^{cxt}(\textbf{h}_t^{enc}, \textbf{h}_{t-1}^{cxt}) \\ \displaybreak[0]
p_{\theta}(\textbf{z}_{t}|\textbf{x}_{\leq t}) = \mathcal{N}(\textbf{z}|\bm{\mu}_t, \bm\sigma_t\textbf{I}) \\
\text{where} \ \ \ \  \bm\mu_t = \text{MLP}_{\theta}(\textbf{h}_{t}^{cxt}) \\
\bm\sigma_t = \text{Softplus}( \ \text{MLP}_{\theta}(\textbf{h}_{t}^{cxt}) \ ) \\ 
\textbf{h}_{t}^{cxt} = [ \textbf{h}_{t}^{cxt} ; \bm{z}_t] \\
\end{gather*}

As a result, our set of transferred parameters contain the additional parameters of MLP$_{\theta}$, included in $\theta_{cxt}^{source}$. \cref{tab:vhred_results} presents the result of using VHRED parameters. Unfortunately, we do not find significant difference between the parameters from VHRED as opposed to HRED. However, the lack of degradation in the results promise possible future improvements in such designs.



\begin{table}
\centering
\makebox[0pt][c]{\parbox{\textwidth}{%
    \begin{minipage}[t]{0.48\hsize}\centering
        
        \resizebox{0.95\linewidth}{!}{
    	\begin{tabular}{|l|c|c|}
    	    \hline
    	    \small{\textbf{Initial Weights}} & \multicolumn{1}{c|}{\textbf{IEMOCAP}} &  \multicolumn{1}{c|}{\textbf{Dailydialog}}\\
    	     &  \small{F-Score} & \small{F-Score} \\
    		\hline
    		\hline
    	    HRED  & 58.5 & 48.0 \\
    	    VHRED  & 58.6 & 48.4 \\
    		\hline
    	\end{tabular}
    	}
    	\caption{\footnotesize{Average performance on ERC with pre-trained weights: $\{\theta^{BERT}+\theta^{cornell}_{cxt}\}$ for VHRED, $\theta^{cornell}_{cxt}$ contain additional parameters modeling the latent prior state.}}
    	\label{tab:vhred_results}

    \end{minipage}
    \hfill
    \begin{minipage}[t]{0.48\hsize}\centering
        
        \resizebox{0.95\linewidth}{!}{
    	\begin{tabular}{|l|c|c|}
    		\hline
    	    \small{\textbf{Generative}} & \multicolumn{1}{c|}{\textbf{IEMOCAP}} &  \multicolumn{1}{c|}{\textbf{Dailydialog}}\\
    	    \small{\textbf{Training}} &  \small{F-Score} & \small{F-Score} \\
    		\hline
    		\hline
    	    Source  & 58.5 & 48.0 \\
    	    Source + Target & 58.0 & 47.2 \\
    		\hline
    	\end{tabular}
    	}
    	\caption{\footnotesize{Average performance on ERC with pre-trained weights: $\{\theta^{BERT}+\theta^{cornell}_{cxt}\}$.}}
    	\label{tab:in-domain_results}
    	
    \end{minipage}
}}
\end{table}

\subsection{In-domain Generative Fine-Tuning}



We try in-domain tuning of the generative HRED model by performing conversation modeling on the ERC resources. Finally, we transfer these re-tuned weights for the discriminative ERC task. However, we do not find this procedure to be helpful ( \cref{tab:in-domain_results} ). TL between generative tasks, especially with small-scale target resources, is a challenging task. As a result, we find sub-optimal generation in ERC datasets whose further transfer for the classification does not provide any improvement.

\subsection{Quality of Generative Models}

Despite their recent developments, generative dialogue models still suffer from numerous shortcomings. Challenges include lack of diversity in the responses, which results in the generation of universal sentences, such as \textit{I don't know}~\cite{li-etal-2016-diversity,DBLP:conf/ijcai/SongLNZZY18}. Coherence in topic and emotions are also difficult to maintain while generating responses~\cite{DBLP:conf/aaai/ZhouHZZL18}. Similar traits are observed in our pre-training experiments. 

Although TL-ERC obtains significant improvement in the results, we obtain it with a simple dialogue model. We, thus, believe that further improvements are possible and is contingent on the quality of the dialogue generator. As research in dialogue systems inch towards the \textit{perfect dialogue generator}, it would also benefit ERC via our proposed TL-ERC framework.

\section{Conclusion} \label{sec:conclusion}

In this paper, we presented a novel framework of transfer learning (TL-ERC) for ERC that uses pre-trained affective information from dialogue generators. We presented various experiments with different scenarios to investigate the effect of this procedure. We found that using such pre-trained weights help the overall task and also provide added benefits in terms of lesser training epochs for good generalization. We primarily experimented on dyadic conversations both in the source and the target tasks. In the future, we aim to investigate the more general setting of multi-party conversations. This setting will increase the complexity of the task, as pre-training would require multi-party data and special training schemes to capture complex influence dynamics.

Code used for this work is publicly available at \url{https://github.com/SenticNet/conv-emotion}.

\section*{Acknowledgement}
This research is supported by Singapore Ministry of Education Academic
Research Fund Tier 1 under MOE's official grant number T1 251RES1820.
We also gratefully acknowledge the support of NVIDIA Corporation with the
donation of a Titan Xp GPU used for this research.

\section*{References}

\bibliographystyle{elsarticle-num-names}
\bibliography{mybibfile}

\begin{thebibliography}{92}
\expandafter\ifx\csname natexlab\endcsname\relax\def\natexlab#1{#1}\fi
\providecommand{\url}[1]{\texttt{#1}}
\providecommand{\href}[2]{#2}
\providecommand{\path}[1]{#1}
\providecommand{\DOIprefix}{doi:}
\providecommand{\ArXivprefix}{arXiv:}
\providecommand{\URLprefix}{URL: }
\providecommand{\Pubmedprefix}{pmid:}
\providecommand{\doi}[1]{\href{http://dx.doi.org/#1}{\path{#1}}}
\providecommand{\Pubmed}[1]{\href{pmid:#1}{\path{#1}}}
\providecommand{\bibinfo}[2]{#2}
\ifx\xfnm\relax \def\xfnm[#1]{\unskip,\space#1}\fi
\bibitem[{Poria et~al.(2019)Poria, Majumder, Mihalcea, and
  Hovy}]{poria2019emotion}
\bibinfo{author}{S.~Poria}, \bibinfo{author}{N.~Majumder},
  \bibinfo{author}{R.~Mihalcea}, \bibinfo{author}{E.~H. Hovy},
\newblock \bibinfo{title}{Emotion recognition in conversation: Research
  challenges, datasets, and recent advances},
\newblock \bibinfo{journal}{{IEEE} Access} \bibinfo{volume}{7}
  (\bibinfo{year}{2019}) \bibinfo{pages}{100943--100953}. \URLprefix
  \url{https://doi.org/10.1109/ACCESS.2019.2929050}.
  \DOIprefix\doi{10.1109/ACCESS.2019.2929050}.
\bibitem[{Chen et~al.(2017)Chen, Liu, Yin, and Tang}]{chen2017survey}
\bibinfo{author}{H.~Chen}, \bibinfo{author}{X.~Liu}, \bibinfo{author}{D.~Yin},
  \bibinfo{author}{J.~Tang},
\newblock \bibinfo{title}{A survey on dialogue systems: Recent advances and new
  frontiers},
\newblock \bibinfo{journal}{{SIGKDD} Explorations} \bibinfo{volume}{19}
  (\bibinfo{year}{2017}) \bibinfo{pages}{25--35}. \URLprefix
  \url{https://doi.org/10.1145/3166054.3166058}.
  \DOIprefix\doi{10.1145/3166054.3166058}.
\bibitem[{Hazarika et~al.(2018)Hazarika, Poria, Zadeh, Cambria, Morency, and
  Zimmermann}]{DBLP:conf/naacl/HazarikaPZCMZ18}
\bibinfo{author}{D.~Hazarika}, \bibinfo{author}{S.~Poria},
  \bibinfo{author}{A.~Zadeh}, \bibinfo{author}{E.~Cambria},
  \bibinfo{author}{L.~Morency}, \bibinfo{author}{R.~Zimmermann},
\newblock \bibinfo{title}{Conversational memory network for emotion recognition
  in dyadic dialogue videos},
\newblock in: \bibinfo{booktitle}{Proceedings of the 2018 Conference of the
  North American Chapter of the Association for Computational Linguistics:
  Human Language Technologies, {NAACL-HLT} 2018, New Orleans, Louisiana, USA,
  June 1-6, 2018, Volume 1 (Long Papers)}, \bibinfo{year}{2018}, pp.
  \bibinfo{pages}{2122--2132}. \URLprefix
  \url{https://www.aclweb.org/anthology/N18-1193/}.
\bibitem[{Pan and Yang(2010)}]{pan2010survey}
\bibinfo{author}{S.~J. Pan}, \bibinfo{author}{Q.~Yang},
\newblock \bibinfo{title}{A survey on transfer learning},
\newblock \bibinfo{journal}{{IEEE} Trans. Knowl. Data Eng.}
  \bibinfo{volume}{22} (\bibinfo{year}{2010}) \bibinfo{pages}{1345--1359}.
  \URLprefix \url{https://doi.org/10.1109/TKDE.2009.191}.
  \DOIprefix\doi{10.1109/TKDE.2009.191}.
\bibitem[{Hovy(1987)}]{hovy1987generating}
\bibinfo{author}{E.~Hovy},
\newblock \bibinfo{title}{Generating natural language under pragmatic
  constraints},
\newblock \bibinfo{journal}{Journal of Pragmatics} \bibinfo{volume}{11}
  (\bibinfo{year}{1987}) \bibinfo{pages}{689--719}.
\bibitem[{Weigand(2017)}]{weigand2017emotions}
\bibinfo{author}{E.~Weigand},
\newblock \bibinfo{title}{Emotions in dialogue},
\newblock \bibinfo{journal}{Dialoganalyse VI/1: Referate der 6. Arbeitstagung,
  Prag 1996} \bibinfo{volume}{16} (\bibinfo{year}{2017}) \bibinfo{pages}{35}.
\bibitem[{Sidnell and Stivers(2012)}]{sidnell2012handbook}
\bibinfo{author}{J.~Sidnell}, \bibinfo{author}{T.~Stivers}, \bibinfo{title}{The
  handbook of conversation analysis}, volume \bibinfo{volume}{121},
  \bibinfo{publisher}{John Wiley \& Sons}, \bibinfo{year}{2012}.
\bibitem[{Koval and Kuppens(2012)}]{koval2012changing}
\bibinfo{author}{P.~Koval}, \bibinfo{author}{P.~Kuppens},
\newblock \bibinfo{title}{Changing emotion dynamics: individual differences in
  the effect of anticipatory social stress on emotional inertia.},
\newblock \bibinfo{journal}{Emotion} \bibinfo{volume}{12}
  (\bibinfo{year}{2012}) \bibinfo{pages}{256}.
\bibitem[{Navarretta(2016)}]{navarretta2016mirroring}
\bibinfo{author}{C.~Navarretta},
\newblock \bibinfo{title}{Mirroring facial expressions and emotions in dyadic
  conversations},
\newblock in: \bibinfo{booktitle}{Proceedings of the Tenth International
  Conference on Language Resources and Evaluation {LREC} 2016, Portoro{\v{z}},
  Slovenia, May 23-28, 2016.}, \bibinfo{year}{2016}. \URLprefix
  \url{http://www.lrec-conf.org/proceedings/lrec2016/summaries/258.html}.
\bibitem[{Shimizu et~al.(2018)Shimizu, Shimizu, and
  Kobayashi}]{shimizu2018pretraining}
\bibinfo{author}{T.~Shimizu}, \bibinfo{author}{N.~Shimizu},
  \bibinfo{author}{H.~Kobayashi},
\newblock \bibinfo{title}{Pretraining sentiment classifiers with unlabeled
  dialog data},
\newblock in: \bibinfo{editor}{I.~Gurevych}, \bibinfo{editor}{Y.~Miyao} (Eds.),
  \bibinfo{booktitle}{Proceedings of the 56th Annual Meeting of the Association
  for Computational Linguistics, {ACL} 2018, Melbourne, Australia, July 15-20,
  2018, Volume 2: Short Papers}, \bibinfo{publisher}{Association for
  Computational Linguistics}, \bibinfo{year}{2018}, pp.
  \bibinfo{pages}{764--770}. \URLprefix
  \url{https://www.aclweb.org/anthology/P18-2121/}.
  \DOIprefix\doi{10.18653/v1/P18-2121}.
\bibitem[{Danescu-Niculescu-Mizil and Lee(2011)}]{danescu2011chameleons}
\bibinfo{author}{C.~Danescu-Niculescu-Mizil}, \bibinfo{author}{L.~Lee},
\newblock \bibinfo{title}{Chameleons in imagined conversations: A new approach
  to understanding coordination of linguistic style in dialogs},
\newblock in: \bibinfo{booktitle}{Proceedings of the 2nd Workshop on Cognitive
  Modeling and Computational Linguistics}, \bibinfo{organization}{Association
  for Computational Linguistics}, \bibinfo{year}{2011}, pp.
  \bibinfo{pages}{76--87}.
\bibitem[{Devlin et~al.(2019)Devlin, Chang, Lee, and
  Toutanova}]{devlin2018bert}
\bibinfo{author}{J.~Devlin}, \bibinfo{author}{M.~Chang},
  \bibinfo{author}{K.~Lee}, \bibinfo{author}{K.~Toutanova},
\newblock \bibinfo{title}{{BERT:} pre-training of deep bidirectional
  transformers for language understanding},
\newblock in:  \cite{DBLP:conf/naacl/2019-1}, \bibinfo{year}{2019}, pp.
  \bibinfo{pages}{4171--4186}. \URLprefix
  \url{https://www.aclweb.org/anthology/N19-1423/}.
\bibitem[{Ando and Zhang(2005)}]{DBLP:journals/jmlr/AndoZ05}
\bibinfo{author}{R.~K. Ando}, \bibinfo{author}{T.~Zhang},
\newblock \bibinfo{title}{A framework for learning predictive structures from
  multiple tasks and unlabeled data},
\newblock \bibinfo{journal}{J. Mach. Learn. Res.} \bibinfo{volume}{6}
  (\bibinfo{year}{2005}) \bibinfo{pages}{1817--1853}. \URLprefix
  \url{http://jmlr.org/papers/v6/ando05a.html}.
\bibitem[{Ruder et~al.(2019)Ruder, Peters, Swayamdipta, and
  Wolf}]{DBLP:conf/naacl/RuderPSW19}
\bibinfo{author}{S.~Ruder}, \bibinfo{author}{M.~E. Peters},
  \bibinfo{author}{S.~Swayamdipta}, \bibinfo{author}{T.~Wolf},
\newblock \bibinfo{title}{Transfer learning in natural language processing},
\newblock in: \bibinfo{booktitle}{Proceedings of the 2019 Conference of the
  North American Chapter of the Association for Computational Linguistics:
  Human Language Technologies, {NAACL-HLT} 2019, Minneapolis, MN, USA, June 2,
  2019, Tutorial Abstracts}, \bibinfo{year}{2019}, pp. \bibinfo{pages}{15--18}.
  \URLprefix \url{https://www.aclweb.org/anthology/N19-5004/}.
\bibitem[{Mikolov et~al.(2013)Mikolov, Sutskever, Chen, Corrado, and
  Dean}]{mikolov2013distributed}
\bibinfo{author}{T.~Mikolov}, \bibinfo{author}{I.~Sutskever},
  \bibinfo{author}{K.~Chen}, \bibinfo{author}{G.~S. Corrado},
  \bibinfo{author}{J.~Dean},
\newblock \bibinfo{title}{Distributed representations of words and phrases and
  their compositionality},
\newblock in: \bibinfo{booktitle}{Advances in Neural Information Processing
  Systems 26: 27th Annual Conference on Neural Information Processing Systems
  2013. Proceedings of a meeting held December 5-8, 2013, Lake Tahoe, Nevada,
  United States.}, \bibinfo{year}{2013}, pp. \bibinfo{pages}{3111--3119}.
\bibitem[{McCann et~al.(2017)McCann, Bradbury, Xiong, and
  Socher}]{mccann2017learned}
\bibinfo{author}{B.~McCann}, \bibinfo{author}{J.~Bradbury},
  \bibinfo{author}{C.~Xiong}, \bibinfo{author}{R.~Socher},
\newblock \bibinfo{title}{Learned in translation: Contextualized word vectors},
\newblock in: \bibinfo{editor}{I.~Guyon}, \bibinfo{editor}{U.~von Luxburg},
  \bibinfo{editor}{S.~Bengio}, \bibinfo{editor}{H.~M. Wallach},
  \bibinfo{editor}{R.~Fergus}, \bibinfo{editor}{S.~V.~N. Vishwanathan},
  \bibinfo{editor}{R.~Garnett} (Eds.), \bibinfo{booktitle}{Advances in Neural
  Information Processing Systems 30: Annual Conference on Neural Information
  Processing Systems 2017, 4-9 December 2017, Long Beach, CA, {USA}},
  \bibinfo{year}{2017}, pp. \bibinfo{pages}{6294--6305}. \URLprefix
  \url{http://papers.nips.cc/paper/7209-learned-in-translation-contextualized-word-vectors}.
\bibitem[{Peters et~al.(2018)Peters, Neumann, Iyyer, Gardner, Clark, Lee, and
  Zettlemoyer}]{peters2018deep}
\bibinfo{author}{M.~E. Peters}, \bibinfo{author}{M.~Neumann},
  \bibinfo{author}{M.~Iyyer}, \bibinfo{author}{M.~Gardner},
  \bibinfo{author}{C.~Clark}, \bibinfo{author}{K.~Lee},
  \bibinfo{author}{L.~Zettlemoyer},
\newblock \bibinfo{title}{Deep contextualized word representations},
\newblock in: \bibinfo{editor}{M.~A. Walker}, \bibinfo{editor}{H.~Ji},
  \bibinfo{editor}{A.~Stent} (Eds.), \bibinfo{booktitle}{Proceedings of the
  2018 Conference of the North American Chapter of the Association for
  Computational Linguistics: Human Language Technologies, {NAACL-HLT} 2018, New
  Orleans, Louisiana, USA, June 1-6, 2018, Volume 1 (Long Papers)},
  \bibinfo{publisher}{Association for Computational Linguistics},
  \bibinfo{year}{2018}, pp. \bibinfo{pages}{2227--2237}. \URLprefix
  \url{https://www.aclweb.org/anthology/N18-1202/}.
\bibitem[{Dai and Le(2015)}]{dai2015semi}
\bibinfo{author}{A.~M. Dai}, \bibinfo{author}{Q.~V. Le},
\newblock \bibinfo{title}{Semi-supervised sequence learning},
\newblock in: \bibinfo{booktitle}{Advances in Neural Information Processing
  Systems 28: Annual Conference on Neural Information Processing Systems 2015,
  December 7-12, 2015, Montreal, Quebec, Canada}, \bibinfo{year}{2015}, pp.
  \bibinfo{pages}{3079--3087}. \URLprefix
  \url{http://papers.nips.cc/paper/5949-semi-supervised-sequence-learning}.
\bibitem[{Yang et~al.(2019)Yang, Dai, Yang, Carbonell, Salakhutdinov, and
  Le}]{DBLP:journals/corr/abs-1906-08237}
\bibinfo{author}{Z.~Yang}, \bibinfo{author}{Z.~Dai}, \bibinfo{author}{Y.~Yang},
  \bibinfo{author}{J.~G. Carbonell}, \bibinfo{author}{R.~Salakhutdinov},
  \bibinfo{author}{Q.~V. Le},
\newblock \bibinfo{title}{Xlnet: Generalized autoregressive pretraining for
  language understanding},
\newblock \bibinfo{journal}{CoRR} \bibinfo{volume}{abs/1906.08237}
  (\bibinfo{year}{2019}). \URLprefix \url{http://arxiv.org/abs/1906.08237}.
  \href{http://arxiv.org/abs/1906.08237}{{\tt arXiv:1906.08237}}.
\bibitem[{Howard and Ruder(2018)}]{howard2018universal}
\bibinfo{author}{J.~Howard}, \bibinfo{author}{S.~Ruder},
\newblock \bibinfo{title}{Universal language model fine-tuning for text
  classification},
\newblock in: \bibinfo{editor}{I.~Gurevych}, \bibinfo{editor}{Y.~Miyao} (Eds.),
  \bibinfo{booktitle}{Proceedings of the 56th Annual Meeting of the Association
  for Computational Linguistics, {ACL} 2018, Melbourne, Australia, July 15-20,
  2018, Volume 1: Long Papers}, \bibinfo{publisher}{Association for
  Computational Linguistics}, \bibinfo{year}{2018}, pp.
  \bibinfo{pages}{328--339}. \URLprefix
  \url{https://www.aclweb.org/anthology/P18-1031/}.
  \DOIprefix\doi{10.18653/v1/P18-1031}.
\bibitem[{Liu et~al.(2019)Liu, Ott, Goyal, Du, Joshi, Chen, Levy, Lewis,
  Zettlemoyer, and Stoyanov}]{DBLP:journals/corr/abs-1907-11692}
\bibinfo{author}{Y.~Liu}, \bibinfo{author}{M.~Ott}, \bibinfo{author}{N.~Goyal},
  \bibinfo{author}{J.~Du}, \bibinfo{author}{M.~Joshi},
  \bibinfo{author}{D.~Chen}, \bibinfo{author}{O.~Levy},
  \bibinfo{author}{M.~Lewis}, \bibinfo{author}{L.~Zettlemoyer},
  \bibinfo{author}{V.~Stoyanov},
\newblock \bibinfo{title}{Roberta: {A} robustly optimized {BERT} pretraining
  approach},
\newblock \bibinfo{journal}{CoRR} \bibinfo{volume}{abs/1907.11692}
  (\bibinfo{year}{2019}). \URLprefix \url{http://arxiv.org/abs/1907.11692}.
  \href{http://arxiv.org/abs/1907.11692}{{\tt arXiv:1907.11692}}.
\bibitem[{Chen and Moschitti(2019)}]{chen2019transfer}
\bibinfo{author}{L.~Chen}, \bibinfo{author}{A.~Moschitti},
\newblock \bibinfo{title}{Transfer learning for sequence labeling using source
  model and target data},
\newblock \bibinfo{journal}{arXiv preprint arXiv:1902.05309}
  (\bibinfo{year}{2019}).
\bibitem[{Qiu et~al.(2018)Qiu, Yang, Ji, Zhou, Huang, Chen, Croft, and
  Lin}]{qiu2018transfer}
\bibinfo{author}{M.~Qiu}, \bibinfo{author}{L.~Yang}, \bibinfo{author}{F.~Ji},
  \bibinfo{author}{W.~Zhou}, \bibinfo{author}{J.~Huang},
  \bibinfo{author}{H.~Chen}, \bibinfo{author}{B.~Croft},
  \bibinfo{author}{W.~Lin},
\newblock \bibinfo{title}{Transfer learning for context-aware question matching
  in information-seeking conversations in e-commerce},
\newblock in: \bibinfo{booktitle}{Proceedings of the 56th Annual Meeting of the
  Association for Computational Linguistics (Volume 2: Short Papers)},
  \bibinfo{year}{2018}, pp. \bibinfo{pages}{208--213}.
\bibitem[{Wu et~al.(2020)Wu, Hoi, Socher, and
  Xiong}]{DBLP:journals/corr/abs-2004-06871}
\bibinfo{author}{C.~Wu}, \bibinfo{author}{S.~C.~H. Hoi},
  \bibinfo{author}{R.~Socher}, \bibinfo{author}{C.~Xiong},
\newblock \bibinfo{title}{Tod-bert: Pre-trained natural language understanding
  for task-oriented dialogues},
\newblock \bibinfo{journal}{CoRR} \bibinfo{volume}{abs/2004.06871}
  (\bibinfo{year}{2020}). \URLprefix \url{https://arxiv.org/abs/2004.06871}.
  \href{http://arxiv.org/abs/2004.06871}{{\tt arXiv:2004.06871}}.
\bibitem[{Henderson et~al.(2019)Henderson, Casanueva, Mrksic, Su, Wen, and
  Vulic}]{DBLP:journals/corr/abs-1911-03688}
\bibinfo{author}{M.~Henderson}, \bibinfo{author}{I.~Casanueva},
  \bibinfo{author}{N.~Mrksic}, \bibinfo{author}{P.~Su},
  \bibinfo{author}{T.~Wen}, \bibinfo{author}{I.~Vulic},
\newblock \bibinfo{title}{Convert: Efficient and accurate conversational
  representations from transformers},
\newblock \bibinfo{journal}{CoRR} \bibinfo{volume}{abs/1911.03688}
  (\bibinfo{year}{2019}). \URLprefix \url{http://arxiv.org/abs/1911.03688}.
  \href{http://arxiv.org/abs/1911.03688}{{\tt arXiv:1911.03688}}.
\bibitem[{Lan et~al.(2020)Lan, Chen, Goodman, Gimpel, Sharma, and
  Soricut}]{DBLP:conf/iclr/LanCGGSS20}
\bibinfo{author}{Z.~Lan}, \bibinfo{author}{M.~Chen},
  \bibinfo{author}{S.~Goodman}, \bibinfo{author}{K.~Gimpel},
  \bibinfo{author}{P.~Sharma}, \bibinfo{author}{R.~Soricut},
\newblock \bibinfo{title}{{ALBERT:} {A} lite {BERT} for self-supervised
  learning of language representations},
\newblock in: \bibinfo{booktitle}{8th International Conference on Learning
  Representations, {ICLR} 2020, Addis Ababa, Ethiopia, April 26-30, 2020},
  \bibinfo{publisher}{OpenReview.net}, \bibinfo{year}{2020}. \URLprefix
  \url{https://openreview.net/forum?id=H1eA7AEtvS}.
\bibitem[{Reimers and Gurevych(2019)}]{DBLP:conf/emnlp/ReimersG19}
\bibinfo{author}{N.~Reimers}, \bibinfo{author}{I.~Gurevych},
\newblock \bibinfo{title}{Sentence-bert: Sentence embeddings using siamese
  bert-networks},
\newblock in: \bibinfo{editor}{K.~Inui}, \bibinfo{editor}{J.~Jiang},
  \bibinfo{editor}{V.~Ng}, \bibinfo{editor}{X.~Wan} (Eds.),
  \bibinfo{booktitle}{Proceedings of the 2019 Conference on Empirical Methods
  in Natural Language Processing and the 9th International Joint Conference on
  Natural Language Processing, {EMNLP-IJCNLP} 2019, Hong Kong, China, November
  3-7, 2019}, \bibinfo{publisher}{Association for Computational Linguistics},
  \bibinfo{year}{2019}, pp. \bibinfo{pages}{3980--3990}. \URLprefix
  \url{https://doi.org/10.18653/v1/D19-1410}.
  \DOIprefix\doi{10.18653/v1/D19-1410}.
\bibitem[{Zhang et~al.(2019)Zhang, Sun, Galley, Chen, Brockett, Gao, Gao, Liu,
  and Dolan}]{DBLP:journals/corr/abs-1911-00536}
\bibinfo{author}{Y.~Zhang}, \bibinfo{author}{S.~Sun},
  \bibinfo{author}{M.~Galley}, \bibinfo{author}{Y.~Chen},
  \bibinfo{author}{C.~Brockett}, \bibinfo{author}{X.~Gao},
  \bibinfo{author}{J.~Gao}, \bibinfo{author}{J.~Liu},
  \bibinfo{author}{B.~Dolan},
\newblock \bibinfo{title}{Dialogpt: Large-scale generative pre-training for
  conversational response generation},
\newblock \bibinfo{journal}{CoRR} \bibinfo{volume}{abs/1911.00536}
  (\bibinfo{year}{2019}). \URLprefix \url{http://arxiv.org/abs/1911.00536}.
  \href{http://arxiv.org/abs/1911.00536}{{\tt arXiv:1911.00536}}.
\bibitem[{Radford et~al.(2019)Radford, Wu, Child, Luan, Amodei, and
  Sutskever}]{radfordlanguage}
\bibinfo{author}{A.~Radford}, \bibinfo{author}{J.~Wu},
  \bibinfo{author}{R.~Child}, \bibinfo{author}{D.~Luan},
  \bibinfo{author}{D.~Amodei}, \bibinfo{author}{I.~Sutskever},
\newblock \bibinfo{title}{Language models are unsupervised multitask learners},
\newblock \bibinfo{journal}{OpenAI Blog} \bibinfo{volume}{1}
  (\bibinfo{year}{2019}) \bibinfo{pages}{9}.
\bibitem[{Cambria et~al.(2019)Cambria, Poria, Hussain, and
  Liu}]{DBLP:journals/cim/CambriaPHL19}
\bibinfo{author}{E.~Cambria}, \bibinfo{author}{S.~Poria},
  \bibinfo{author}{A.~Hussain}, \bibinfo{author}{B.~Liu},
\newblock \bibinfo{title}{Computational intelligence for affective computing
  and sentiment analysis [guest editorial]},
\newblock \bibinfo{journal}{{IEEE} Comput. Intell. Mag.} \bibinfo{volume}{14}
  (\bibinfo{year}{2019}) \bibinfo{pages}{16--17}. \URLprefix
  \url{https://doi.org/10.1109/MCI.2019.2901082}.
  \DOIprefix\doi{10.1109/MCI.2019.2901082}.
\bibitem[{Imani and Montazer(2019)}]{DBLP:journals/jnca/ImaniM19}
\bibinfo{author}{M.~Imani}, \bibinfo{author}{G.~A. Montazer},
\newblock \bibinfo{title}{A survey of emotion recognition methods with emphasis
  on e-learning environments},
\newblock \bibinfo{journal}{J. Netw. Comput. Appl.} \bibinfo{volume}{147}
  (\bibinfo{year}{2019}). \URLprefix
  \url{https://doi.org/10.1016/j.jnca.2019.102423}.
  \DOIprefix\doi{10.1016/j.jnca.2019.102423}.
\bibitem[{Liscombe et~al.(2005)Liscombe, Riccardi, and
  Hakkani{-}T{\"{u}}r}]{DBLP:conf/interspeech/LiscombeRH05}
\bibinfo{author}{J.~Liscombe}, \bibinfo{author}{G.~Riccardi},
  \bibinfo{author}{D.~Hakkani{-}T{\"{u}}r},
\newblock \bibinfo{title}{Using context to improve emotion detection in spoken
  dialog systems},
\newblock in: \bibinfo{booktitle}{{INTERSPEECH} 2005 - Eurospeech, 9th European
  Conference on Speech Communication and Technology, Lisbon, Portugal,
  September 4-8, 2005}, \bibinfo{publisher}{{ISCA}}, \bibinfo{year}{2005}, pp.
  \bibinfo{pages}{1845--1848}. \URLprefix
  \url{http://www.isca-speech.org/archive/interspeech\_2005/i05\_1845.html}.
\bibitem[{Schiaffino and Amandi(2009)}]{DBLP:series/lncs/SchiaffinoA09}
\bibinfo{author}{S.~N. Schiaffino}, \bibinfo{author}{A.~Amandi},
\newblock \bibinfo{title}{Intelligent user profiling},
\newblock in: \bibinfo{editor}{M.~Bramer} (Ed.), \bibinfo{booktitle}{Artificial
  Intelligence: An International Perspective}, volume \bibinfo{volume}{5640} of
  \textit{\bibinfo{series}{Lecture Notes in Computer Science}},
  \bibinfo{publisher}{Springer}, \bibinfo{year}{2009}, pp.
  \bibinfo{pages}{193--216}. \URLprefix
  \url{https://doi.org/10.1007/978-3-642-03226-4\_11}.
  \DOIprefix\doi{10.1007/978-3-642-03226-4\_11}.
\bibitem[{Li and Deng(2018)}]{DBLP:journals/corr/abs-1804-08348}
\bibinfo{author}{S.~Li}, \bibinfo{author}{W.~Deng},
\newblock \bibinfo{title}{Deep facial expression recognition: {A} survey},
\newblock \bibinfo{journal}{CoRR} \bibinfo{volume}{abs/1804.08348}
  (\bibinfo{year}{2018}). \URLprefix \url{http://arxiv.org/abs/1804.08348}.
  \href{http://arxiv.org/abs/1804.08348}{{\tt arXiv:1804.08348}}.
\bibitem[{Wang et~al.(2018)Wang, Phillips, Dong, and
  Zhang}]{DBLP:journals/ijon/WangPDZ18}
\bibinfo{author}{S.~Wang}, \bibinfo{author}{P.~Phillips},
  \bibinfo{author}{Z.~Dong}, \bibinfo{author}{Y.~Zhang},
\newblock \bibinfo{title}{Intelligent facial emotion recognition based on
  stationary wavelet entropy and jaya algorithm},
\newblock \bibinfo{journal}{Neurocomputing} \bibinfo{volume}{272}
  (\bibinfo{year}{2018}) \bibinfo{pages}{668--676}. \URLprefix
  \url{https://doi.org/10.1016/j.neucom.2017.08.015}.
  \DOIprefix\doi{10.1016/j.neucom.2017.08.015}.
\bibitem[{Drakopoulos et~al.(2019)Drakopoulos, Pikramenos, Spyrou, and
  Perantonis}]{DBLP:conf/webist/DrakopoulosPSP19}
\bibinfo{author}{G.~Drakopoulos}, \bibinfo{author}{G.~Pikramenos},
  \bibinfo{author}{E.~D. Spyrou}, \bibinfo{author}{S.~J. Perantonis},
\newblock \bibinfo{title}{Emotion recognition from speech: {A} survey},
\newblock in: \bibinfo{editor}{A.~Bozzon}, \bibinfo{editor}{F.~J.~D. Mayo},
  \bibinfo{editor}{J.~Filipe} (Eds.), \bibinfo{booktitle}{Proceedings of the
  15th International Conference on Web Information Systems and Technologies,
  {WEBIST} 2019, Vienna, Austria, September 18-20, 2019},
  \bibinfo{publisher}{ScitePress}, \bibinfo{year}{2019}, pp.
  \bibinfo{pages}{432--439}.
\bibitem[{Anagnostopoulos et~al.(2015)Anagnostopoulos, Iliou, and
  Giannoukos}]{DBLP:journals/air/AnagnostopoulosIG15}
\bibinfo{author}{C.~Anagnostopoulos}, \bibinfo{author}{T.~Iliou},
  \bibinfo{author}{I.~Giannoukos},
\newblock \bibinfo{title}{Features and classifiers for emotion recognition from
  speech: a survey from 2000 to 2011},
\newblock \bibinfo{journal}{Artif. Intell. Rev.} \bibinfo{volume}{43}
  (\bibinfo{year}{2015}) \bibinfo{pages}{155--177}. \URLprefix
  \url{https://doi.org/10.1007/s10462-012-9368-5}.
  \DOIprefix\doi{10.1007/s10462-012-9368-5}.
\bibitem[{Mar{\'{e}}chal et~al.(2019)Mar{\'{e}}chal, Mikolajewski, Tyburek,
  Prokopowicz, Bougueroua, Ancourt, and
  Wegrzyn{-}Wolska}]{DBLP:series/lncs/MarechalMTPBAW19}
\bibinfo{author}{C.~Mar{\'{e}}chal}, \bibinfo{author}{D.~Mikolajewski},
  \bibinfo{author}{K.~Tyburek}, \bibinfo{author}{P.~Prokopowicz},
  \bibinfo{author}{L.~Bougueroua}, \bibinfo{author}{C.~Ancourt},
  \bibinfo{author}{K.~Wegrzyn{-}Wolska},
\newblock \bibinfo{title}{Survey on ai-based multimodal methods for emotion
  detection},
\newblock in: \bibinfo{editor}{J.~Kolodziej},
  \bibinfo{editor}{H.~Gonz{\'{a}}lez{-}V{\'{e}}lez} (Eds.),
  \bibinfo{booktitle}{High-Performance Modelling and Simulation for Big Data
  Applications - Selected Results of the {COST} Action {IC1406} cHiPSet},
  volume \bibinfo{volume}{11400} of \textit{\bibinfo{series}{Lecture Notes in
  Computer Science}}, \bibinfo{publisher}{Springer}, \bibinfo{year}{2019}, pp.
  \bibinfo{pages}{307--324}. \URLprefix
  \url{https://doi.org/10.1007/978-3-030-16272-6\_11}.
  \DOIprefix\doi{10.1007/978-3-030-16272-6\_11}.
\bibitem[{Alm et~al.(2005)Alm, Roth, and Sproat}]{DBLP:conf/naacl/AlmRS05}
\bibinfo{author}{C.~O. Alm}, \bibinfo{author}{D.~Roth},
  \bibinfo{author}{R.~Sproat},
\newblock \bibinfo{title}{Emotions from text: Machine learning for text-based
  emotion prediction},
\newblock in: \bibinfo{booktitle}{{HLT/EMNLP} 2005, Human Language Technology
  Conference and Conference on Empirical Methods in Natural Language
  Processing, Proceedings of the Conference, 6-8 October 2005, Vancouver,
  British Columbia, Canada}, \bibinfo{publisher}{The Association for
  Computational Linguistics}, \bibinfo{year}{2005}, pp.
  \bibinfo{pages}{579--586}. \URLprefix
  \url{https://www.aclweb.org/anthology/H05-1073/}.
\bibitem[{Strapparava and Mihalcea(2008)}]{DBLP:conf/sac/StrapparavaM08}
\bibinfo{author}{C.~Strapparava}, \bibinfo{author}{R.~Mihalcea},
\newblock \bibinfo{title}{Learning to identify emotions in text},
\newblock in: \bibinfo{editor}{R.~L. Wainwright}, \bibinfo{editor}{H.~Haddad}
  (Eds.), \bibinfo{booktitle}{Proceedings of the 2008 {ACM} Symposium on
  Applied Computing (SAC), Fortaleza, Ceara, Brazil, March 16-20, 2008},
  \bibinfo{publisher}{{ACM}}, \bibinfo{year}{2008}, pp.
  \bibinfo{pages}{1556--1560}. \URLprefix
  \url{https://doi.org/10.1145/1363686.1364052}.
  \DOIprefix\doi{10.1145/1363686.1364052}.
\bibitem[{Strapparava and Valitutti(2004)}]{DBLP:conf/lrec/StrapparavaV04}
\bibinfo{author}{C.~Strapparava}, \bibinfo{author}{A.~Valitutti},
\newblock \bibinfo{title}{Wordnet affect: an affective extension of wordnet},
\newblock in: \bibinfo{booktitle}{Proceedings of the Fourth International
  Conference on Language Resources and Evaluation, {LREC} 2004, May 26-28,
  2004, Lisbon, Portugal}, \bibinfo{publisher}{European Language Resources
  Association}, \bibinfo{year}{2004}. \URLprefix
  \url{http://www.lrec-conf.org/proceedings/lrec2004/summaries/369.htm}.
\bibitem[{Esuli and Sebastiani(2006)}]{DBLP:conf/lrec/Esuli006}
\bibinfo{author}{A.~Esuli}, \bibinfo{author}{F.~Sebastiani},
\newblock \bibinfo{title}{{SENTIWORDNET:} {A} publicly available lexical
  resource for opinion mining},
\newblock in: \bibinfo{editor}{N.~Calzolari}, \bibinfo{editor}{K.~Choukri},
  \bibinfo{editor}{A.~Gangemi}, \bibinfo{editor}{B.~Maegaard},
  \bibinfo{editor}{J.~Mariani}, \bibinfo{editor}{J.~Odijk},
  \bibinfo{editor}{D.~Tapias} (Eds.), \bibinfo{booktitle}{Proceedings of the
  Fifth International Conference on Language Resources and Evaluation, {LREC}
  2006, Genoa, Italy, May 22-28, 2006}, \bibinfo{publisher}{European Language
  Resources Association {(ELRA)}}, \bibinfo{year}{2006}, pp.
  \bibinfo{pages}{417--422}. \URLprefix
  \url{http://www.lrec-conf.org/proceedings/lrec2006/summaries/384.html}.
\bibitem[{Wang et~al.(2012)Wang, Chen, Thirunarayan, and
  Sheth}]{DBLP:conf/socialcom/0002CTS12}
\bibinfo{author}{W.~Wang}, \bibinfo{author}{L.~Chen},
  \bibinfo{author}{K.~Thirunarayan}, \bibinfo{author}{A.~P. Sheth},
\newblock \bibinfo{title}{Harnessing twitter "big data" for automatic emotion
  identification},
\newblock in: \bibinfo{booktitle}{2012 International Conference on Privacy,
  Security, Risk and Trust, {PASSAT} 2012, and 2012 International Confernece on
  Social Computing, SocialCom 2012, Amsterdam, Netherlands, September 3-5,
  2012}, \bibinfo{publisher}{{IEEE} Computer Society}, \bibinfo{year}{2012},
  pp. \bibinfo{pages}{587--592}. \URLprefix
  \url{https://doi.org/10.1109/SocialCom-PASSAT.2012.119}.
  \DOIprefix\doi{10.1109/SocialCom-PASSAT.2012.119}.
\bibitem[{Choi et~al.(2018)Choi, Song, and Lee}]{choi-etal-2018-convolutional}
\bibinfo{author}{W.~Y. Choi}, \bibinfo{author}{K.~Y. Song},
  \bibinfo{author}{C.~W. Lee},
\newblock \bibinfo{title}{Convolutional attention networks for multimodal
  emotion recognition from speech and text data},
\newblock in: \bibinfo{booktitle}{Proceedings of Grand Challenge and Workshop
  on Human Multimodal Language (Challenge-{HML})},
  \bibinfo{publisher}{Association for Computational Linguistics},
  \bibinfo{address}{Melbourne, Australia}, \bibinfo{year}{2018}, pp.
  \bibinfo{pages}{28--34}. \URLprefix
  \url{https://www.aclweb.org/anthology/W18-3304}.
  \DOIprefix\doi{10.18653/v1/W18-3304}.
\bibitem[{Chernykh et~al.(2017)Chernykh, Sterling, and
  Prihodko}]{DBLP:journals/corr/ChernykhSP17}
\bibinfo{author}{V.~Chernykh}, \bibinfo{author}{G.~Sterling},
  \bibinfo{author}{P.~Prihodko},
\newblock \bibinfo{title}{Emotion recognition from speech with recurrent neural
  networks},
\newblock \bibinfo{journal}{CoRR} \bibinfo{volume}{abs/1701.08071}
  (\bibinfo{year}{2017}). \URLprefix \url{http://arxiv.org/abs/1701.08071}.
  \href{http://arxiv.org/abs/1701.08071}{{\tt arXiv:1701.08071}}.
\bibitem[{Mirsamadi et~al.(2017)Mirsamadi, Barsoum, and
  Zhang}]{DBLP:conf/icassp/MirsamadiBZ17}
\bibinfo{author}{S.~Mirsamadi}, \bibinfo{author}{E.~Barsoum},
  \bibinfo{author}{C.~Zhang},
\newblock \bibinfo{title}{Automatic speech emotion recognition using recurrent
  neural networks with local attention},
\newblock in: \bibinfo{booktitle}{2017 {IEEE} International Conference on
  Acoustics, Speech and Signal Processing, {ICASSP} 2017, New Orleans, LA, USA,
  March 5-9, 2017}, \bibinfo{publisher}{{IEEE}}, \bibinfo{year}{2017}, pp.
  \bibinfo{pages}{2227--2231}. \URLprefix
  \url{https://doi.org/10.1109/ICASSP.2017.7952552}.
  \DOIprefix\doi{10.1109/ICASSP.2017.7952552}.
\bibitem[{Dragoni(2019)}]{DBLP:journals/cim/Dragoni19}
\bibinfo{author}{M.~Dragoni},
\newblock \bibinfo{title}{An evolutionary strategy for concept-based
  multi-domain sentiment analysis},
\newblock \bibinfo{journal}{{IEEE} Comput. Intell. Mag.} \bibinfo{volume}{14}
  (\bibinfo{year}{2019}) \bibinfo{pages}{18--27}. \URLprefix
  \url{https://doi.org/10.1109/MCI.2019.2901083}.
  \DOIprefix\doi{10.1109/MCI.2019.2901083}.
\bibitem[{Han et~al.(2019)Han, Zhang, and Schuller}]{DBLP:journals/cim/HanZS19}
\bibinfo{author}{J.~Han}, \bibinfo{author}{Z.~Zhang}, \bibinfo{author}{B.~W.
  Schuller},
\newblock \bibinfo{title}{Adversarial training in affective computing and
  sentiment analysis: Recent advances and perspectives [review article]},
\newblock \bibinfo{journal}{{IEEE} Comput. Intell. Mag.} \bibinfo{volume}{14}
  (\bibinfo{year}{2019}) \bibinfo{pages}{68--81}. \URLprefix
  \url{https://doi.org/10.1109/MCI.2019.2901088}.
  \DOIprefix\doi{10.1109/MCI.2019.2901088}.
\bibitem[{Yu et~al.(2018)Yu, Marujo, Jiang, Karuturi, and
  Brendel}]{yu2018improving}
\bibinfo{author}{J.~Yu}, \bibinfo{author}{L.~Marujo},
  \bibinfo{author}{J.~Jiang}, \bibinfo{author}{P.~Karuturi},
  \bibinfo{author}{W.~Brendel},
\newblock \bibinfo{title}{Improving multi-label emotion classification via
  sentiment classification with dual attention transfer network},
\newblock in:  \cite{DBLP:conf/emnlp/2018}, \bibinfo{year}{2018}, pp.
  \bibinfo{pages}{1097--1102}. \URLprefix
  \url{https://www.aclweb.org/anthology/D18-1137/}.
\bibitem[{Daval{-}Frerot et~al.(2018)Daval{-}Frerot, Bouchekif, and
  Moreau}]{DBLP:conf/semeval/Daval-FrerotBM18}
\bibinfo{author}{G.~Daval{-}Frerot}, \bibinfo{author}{A.~Bouchekif},
  \bibinfo{author}{A.~Moreau},
\newblock \bibinfo{title}{Epita at semeval-2018 task 1: Sentiment analysis
  using transfer learning approach},
\newblock in: \bibinfo{booktitle}{Proceedings of The 12th International
  Workshop on Semantic Evaluation, SemEval@NAACL-HLT 2018, New Orleans,
  Louisiana, USA, June 5-6, 2018}, \bibinfo{year}{2018}, pp.
  \bibinfo{pages}{151--155}. \URLprefix
  \url{https://www.aclweb.org/anthology/S18-1021/}.
\bibitem[{Bouchekif et~al.(2019)Bouchekif, Joshi, Bouchekif, and
  Afli}]{bouchekif2019epita}
\bibinfo{author}{A.~Bouchekif}, \bibinfo{author}{P.~Joshi},
  \bibinfo{author}{L.~Bouchekif}, \bibinfo{author}{H.~Afli},
\newblock \bibinfo{title}{Epita-adapt at semeval-2019 task 3: Detecting
  emotions in textual conversations using deep learning models combination},
\newblock in: \bibinfo{booktitle}{Proceedings of the 13th International
  Workshop on Semantic Evaluation}, \bibinfo{year}{2019}, pp.
  \bibinfo{pages}{215--219}.
\bibitem[{Ng et~al.(2015)Ng, Nguyen, Vonikakis, and Winkler}]{ng2015deep}
\bibinfo{author}{H.~Ng}, \bibinfo{author}{V.~D. Nguyen},
  \bibinfo{author}{V.~Vonikakis}, \bibinfo{author}{S.~Winkler},
\newblock \bibinfo{title}{Deep learning for emotion recognition on small
  datasets using transfer learning},
\newblock in: \bibinfo{editor}{Z.~Zhang}, \bibinfo{editor}{P.~Cohen},
  \bibinfo{editor}{D.~Bohus}, \bibinfo{editor}{R.~Horaud},
  \bibinfo{editor}{H.~Meng} (Eds.), \bibinfo{booktitle}{Proceedings of the 2015
  {ACM} on International Conference on Multimodal Interaction, Seattle, WA,
  USA, November 09 - 13, 2015}, \bibinfo{publisher}{{ACM}},
  \bibinfo{year}{2015}, pp. \bibinfo{pages}{443--449}. \URLprefix
  \url{http://doi.acm.org/10.1145/2818346.2830593}.
  \DOIprefix\doi{10.1145/2818346.2830593}.
\bibitem[{Deng et~al.(2013)Deng, Zhang, Marchi, and Schuller}]{deng2013sparse}
\bibinfo{author}{J.~Deng}, \bibinfo{author}{Z.~Zhang},
  \bibinfo{author}{E.~Marchi}, \bibinfo{author}{B.~W. Schuller},
\newblock \bibinfo{title}{Sparse autoencoder-based feature transfer learning
  for speech emotion recognition},
\newblock in: \bibinfo{booktitle}{2013 Humaine Association Conference on
  Affective Computing and Intelligent Interaction, {ACII} 2013, Geneva,
  Switzerland, September 2-5, 2013}, \bibinfo{publisher}{{IEEE} Computer
  Society}, \bibinfo{year}{2013}, pp. \bibinfo{pages}{511--516}. \URLprefix
  \url{https://doi.org/10.1109/ACII.2013.90}.
  \DOIprefix\doi{10.1109/ACII.2013.90}.
\bibitem[{Felbo et~al.(2017)Felbo, Mislove, S{\o}gaard, Rahwan, and
  Lehmann}]{DBLP:conf/emnlp/FelboMSRL17}
\bibinfo{author}{B.~Felbo}, \bibinfo{author}{A.~Mislove},
  \bibinfo{author}{A.~S{\o}gaard}, \bibinfo{author}{I.~Rahwan},
  \bibinfo{author}{S.~Lehmann},
\newblock \bibinfo{title}{Using millions of emoji occurrences to learn
  any-domain representations for detecting sentiment, emotion and sarcasm},
\newblock in: \bibinfo{editor}{M.~Palmer}, \bibinfo{editor}{R.~Hwa},
  \bibinfo{editor}{S.~Riedel} (Eds.), \bibinfo{booktitle}{Proceedings of the
  2017 Conference on Empirical Methods in Natural Language Processing, {EMNLP}
  2017, Copenhagen, Denmark, September 9-11, 2017},
  \bibinfo{publisher}{Association for Computational Linguistics},
  \bibinfo{year}{2017}, pp. \bibinfo{pages}{1615--1625}. \URLprefix
  \url{https://doi.org/10.18653/v1/d17-1169}.
  \DOIprefix\doi{10.18653/v1/d17-1169}.
\bibitem[{Gonz{\'{a}}lez{-}Gardu{\~{n}}o
  et~al.(2019)Gonz{\'{a}}lez{-}Gardu{\~{n}}o, Hansen, Bingel, Augenstein, and
  S{\o}gaard}]{DBLP:conf/semeval/Gonzalez-Garduno19}
\bibinfo{author}{A.~V. Gonz{\'{a}}lez{-}Gardu{\~{n}}o},
  \bibinfo{author}{V.~P.~B. Hansen}, \bibinfo{author}{J.~Bingel},
  \bibinfo{author}{I.~Augenstein}, \bibinfo{author}{A.~S{\o}gaard},
\newblock \bibinfo{title}{Coastal at semeval-2019 task 3: Affect classification
  in dialogue using attentive bilstms},
\newblock in: \bibinfo{booktitle}{Proceedings of the 13th International
  Workshop on Semantic Evaluation, SemEval@NAACL-HLT 2019, Minneapolis, MN,
  USA, June 6-7, 2019}, \bibinfo{year}{2019}, pp. \bibinfo{pages}{169--174}.
  \URLprefix \url{https://www.aclweb.org/anthology/S19-2026/}.
\bibitem[{Jiao et~al.(2019)Jiao, Yang, King, and
  Lyu}]{DBLP:conf/naacl/JiaoYKL19}
\bibinfo{author}{W.~Jiao}, \bibinfo{author}{H.~Yang},
  \bibinfo{author}{I.~King}, \bibinfo{author}{M.~R. Lyu},
\newblock \bibinfo{title}{Higru: Hierarchical gated recurrent units for
  utterance-level emotion recognition},
\newblock in:  \cite{DBLP:conf/naacl/2019-1}, \bibinfo{year}{2019}, pp.
  \bibinfo{pages}{397--406}. \URLprefix
  \url{https://www.aclweb.org/anthology/N19-1037/}.
\bibitem[{Hazarika et~al.(2018)Hazarika, Poria, Mihalcea, Cambria, and
  Zimmermann}]{hazarika2018icon}
\bibinfo{author}{D.~Hazarika}, \bibinfo{author}{S.~Poria},
  \bibinfo{author}{R.~Mihalcea}, \bibinfo{author}{E.~Cambria},
  \bibinfo{author}{R.~Zimmermann},
\newblock \bibinfo{title}{{ICON:} interactive conversational memory network for
  multimodal emotion detection},
\newblock in:  \cite{DBLP:conf/emnlp/2018}, \bibinfo{year}{2018}, pp.
  \bibinfo{pages}{2594--2604}. \URLprefix
  \url{https://www.aclweb.org/anthology/D18-1280/}.
\bibitem[{Ghosal et~al.(2019)Ghosal, Majumder, Poria, Chhaya, and
  Gelbukh}]{DBLP:journals/corr/abs-1908-11540}
\bibinfo{author}{D.~Ghosal}, \bibinfo{author}{N.~Majumder},
  \bibinfo{author}{S.~Poria}, \bibinfo{author}{N.~Chhaya},
  \bibinfo{author}{A.~F. Gelbukh},
\newblock \bibinfo{title}{Dialoguegcn: {A} graph convolutional neural network
  for emotion recognition in conversation},
\newblock \bibinfo{journal}{CoRR} \bibinfo{volume}{abs/1908.11540}
  (\bibinfo{year}{2019}). \URLprefix \url{http://arxiv.org/abs/1908.11540}.
  \href{http://arxiv.org/abs/1908.11540}{{\tt arXiv:1908.11540}}.
\bibitem[{Zhang et~al.(2019{\natexlab{a}})Zhang, Wu, Sun, Li, Zhu, and
  Zhou}]{DBLP:conf/ijcai/ZhangWSLZZ19}
\bibinfo{author}{D.~Zhang}, \bibinfo{author}{L.~Wu}, \bibinfo{author}{C.~Sun},
  \bibinfo{author}{S.~Li}, \bibinfo{author}{Q.~Zhu}, \bibinfo{author}{G.~Zhou},
\newblock \bibinfo{title}{Modeling both context- and speaker-sensitive
  dependence for emotion detection in multi-speaker conversations},
\newblock in:  \cite{DBLP:conf/ijcai/2019}, \bibinfo{year}{2019}{\natexlab{a}},
  pp. \bibinfo{pages}{5415--5421}. \URLprefix
  \url{https://doi.org/10.24963/ijcai.2019/752}.
  \DOIprefix\doi{10.24963/ijcai.2019/752}.
\bibitem[{Zhang et~al.(2019{\natexlab{b}})Zhang, Li, Song, Zhang, and
  Wang}]{DBLP:conf/ijcai/ZhangL0ZW19}
\bibinfo{author}{Y.~Zhang}, \bibinfo{author}{Q.~Li}, \bibinfo{author}{D.~Song},
  \bibinfo{author}{P.~Zhang}, \bibinfo{author}{P.~Wang},
\newblock \bibinfo{title}{Quantum-inspired interactive networks for
  conversational sentiment analysis},
\newblock in:  \cite{DBLP:conf/ijcai/2019}, \bibinfo{year}{2019}{\natexlab{b}},
  pp. \bibinfo{pages}{5436--5442}. \URLprefix
  \url{https://doi.org/10.24963/ijcai.2019/755}.
  \DOIprefix\doi{10.24963/ijcai.2019/755}.
\bibitem[{Majumder et~al.(2019)Majumder, Poria, Hazarika, Mihalcea, Gelbukh,
  and Cambria}]{DBLP:conf/aaai/MajumderPHMGC19}
\bibinfo{author}{N.~Majumder}, \bibinfo{author}{S.~Poria},
  \bibinfo{author}{D.~Hazarika}, \bibinfo{author}{R.~Mihalcea},
  \bibinfo{author}{A.~F. Gelbukh}, \bibinfo{author}{E.~Cambria},
\newblock \bibinfo{title}{Dialoguernn: An attentive {RNN} for emotion detection
  in conversations},
\newblock in: \bibinfo{booktitle}{The Thirty-Third {AAAI} Conference on
  Artificial Intelligence, {AAAI} 2019, The Thirty-First Innovative
  Applications of Artificial Intelligence Conference, {IAAI} 2019, The Ninth
  {AAAI} Symposium on Educational Advances in Artificial Intelligence, {EAAI}
  2019, Honolulu, Hawaii, USA, January 27 - February 1, 2019.},
  \bibinfo{publisher}{{AAAI} Press}, \bibinfo{year}{2019}, pp.
  \bibinfo{pages}{6818--6825}. \URLprefix
  \url{https://aaai.org/ojs/index.php/AAAI/article/view/4657}.
\bibitem[{Zhong et~al.(2019)Zhong, Wang, and
  Miao}]{DBLP:journals/corr/abs-1909-10681}
\bibinfo{author}{P.~Zhong}, \bibinfo{author}{D.~Wang},
  \bibinfo{author}{C.~Miao},
\newblock \bibinfo{title}{Knowledge-enriched transformer for emotion detection
  in textual conversations},
\newblock \bibinfo{journal}{CoRR} \bibinfo{volume}{abs/1909.10681}
  (\bibinfo{year}{2019}). \URLprefix \url{http://arxiv.org/abs/1909.10681}.
  \href{http://arxiv.org/abs/1909.10681}{{\tt arXiv:1909.10681}}.
\bibitem[{Chatterjee et~al.(2019)Chatterjee, Narahari, Joshi, and
  Agrawal}]{DBLP:conf/semeval/ChatterjeeNJA19}
\bibinfo{author}{A.~Chatterjee}, \bibinfo{author}{K.~N. Narahari},
  \bibinfo{author}{M.~Joshi}, \bibinfo{author}{P.~Agrawal},
\newblock \bibinfo{title}{Semeval-2019 task 3: Emocontext contextual emotion
  detection in text},
\newblock in: \bibinfo{editor}{J.~May}, \bibinfo{editor}{E.~Shutova},
  \bibinfo{editor}{A.~Herbelot}, \bibinfo{editor}{X.~Zhu},
  \bibinfo{editor}{M.~Apidianaki}, \bibinfo{editor}{S.~M. Mohammad} (Eds.),
  \bibinfo{booktitle}{Proceedings of the 13th International Workshop on
  Semantic Evaluation, SemEval@NAACL-HLT 2019, Minneapolis, MN, USA, June 6-7,
  2019}, \bibinfo{publisher}{Association for Computational Linguistics},
  \bibinfo{year}{2019}, pp. \bibinfo{pages}{39--48}. \URLprefix
  \url{https://doi.org/10.18653/v1/s19-2005}.
  \DOIprefix\doi{10.18653/v1/s19-2005}.
\bibitem[{Huang et~al.(2019)Huang, Lee, Ma, Chen, Yu, and
  Chen}]{DBLP:journals/corr/abs-1908-06264}
\bibinfo{author}{Y.~Huang}, \bibinfo{author}{S.~Lee}, \bibinfo{author}{M.~Ma},
  \bibinfo{author}{Y.~Chen}, \bibinfo{author}{Y.~Yu},
  \bibinfo{author}{Y.~Chen},
\newblock \bibinfo{title}{Emotionx-idea: Emotion {BERT} - an affectional model
  for conversation},
\newblock \bibinfo{journal}{CoRR} \bibinfo{volume}{abs/1908.06264}
  (\bibinfo{year}{2019}). \URLprefix \url{http://arxiv.org/abs/1908.06264}.
  \href{http://arxiv.org/abs/1908.06264}{{\tt arXiv:1908.06264}}.
\bibitem[{Jiao et~al.(2019)Jiao, Lyu, and
  King}]{DBLP:journals/corr/abs-1910-08916}
\bibinfo{author}{W.~Jiao}, \bibinfo{author}{M.~R. Lyu},
  \bibinfo{author}{I.~King},
\newblock \bibinfo{title}{Pt-code: Pre-trained context-dependent encoder for
  utterance-level emotion recognition},
\newblock \bibinfo{journal}{CoRR} \bibinfo{volume}{abs/1910.08916}
  (\bibinfo{year}{2019}). \URLprefix \url{http://arxiv.org/abs/1910.08916}.
  \href{http://arxiv.org/abs/1910.08916}{{\tt arXiv:1910.08916}}.
\bibitem[{Serban et~al.(2016)Serban, Sordoni, Bengio, Courville, and
  Pineau}]{serban2016building}
\bibinfo{author}{I.~V. Serban}, \bibinfo{author}{A.~Sordoni},
  \bibinfo{author}{Y.~Bengio}, \bibinfo{author}{A.~C. Courville},
  \bibinfo{author}{J.~Pineau},
\newblock \bibinfo{title}{Building end-to-end dialogue systems using generative
  hierarchical neural network models},
\newblock in: \bibinfo{editor}{D.~Schuurmans}, \bibinfo{editor}{M.~P. Wellman}
  (Eds.), \bibinfo{booktitle}{Proceedings of the Thirtieth {AAAI} Conference on
  Artificial Intelligence, February 12-17, 2016, Phoenix, Arizona, {USA.}},
  \bibinfo{publisher}{{AAAI} Press}, \bibinfo{year}{2016}, pp.
  \bibinfo{pages}{3776--3784}. \URLprefix
  \url{http://www.aaai.org/ocs/index.php/AAAI/AAAI16/paper/view/11957}.
\bibitem[{Cho et~al.(2014)Cho, van Merrienboer, Gulcehre, Bahdanau, Bougares,
  Schwenk, and Bengio}]{cho2014learning}
\bibinfo{author}{K.~Cho}, \bibinfo{author}{B.~van Merrienboer},
  \bibinfo{author}{C.~Gulcehre}, \bibinfo{author}{D.~Bahdanau},
  \bibinfo{author}{F.~Bougares}, \bibinfo{author}{H.~Schwenk},
  \bibinfo{author}{Y.~Bengio},
\newblock \bibinfo{title}{Learning phrase representations using rnn
  encoder--decoder for statistical machine translation},
\newblock in: \bibinfo{booktitle}{Proceedings of the 2014 Conference on
  Empirical Methods in Natural Language Processing (EMNLP)},
  \bibinfo{year}{2014}, pp. \bibinfo{pages}{1724--1734}.
\bibitem[{Poria et~al.(2017)Poria, Cambria, Hazarika, Majumder, Zadeh, and
  Morency}]{poria2017context}
\bibinfo{author}{S.~Poria}, \bibinfo{author}{E.~Cambria},
  \bibinfo{author}{D.~Hazarika}, \bibinfo{author}{N.~Majumder},
  \bibinfo{author}{A.~Zadeh}, \bibinfo{author}{L.~Morency},
\newblock \bibinfo{title}{Context-dependent sentiment analysis in
  user-generated videos},
\newblock in: \bibinfo{booktitle}{Proceedings of the 55th Annual Meeting of the
  Association for Computational Linguistics, {ACL} 2017, Vancouver, Canada,
  July 30 - August 4, Volume 1: Long Papers}, \bibinfo{year}{2017}, pp.
  \bibinfo{pages}{873--883}. \URLprefix
  \url{https://doi.org/10.18653/v1/P17-1081}.
  \DOIprefix\doi{10.18653/v1/P17-1081}.
\bibitem[{Lowe et~al.(2015)Lowe, Pow, Serban, and Pineau}]{lowe2015ubuntu}
\bibinfo{author}{R.~Lowe}, \bibinfo{author}{N.~Pow},
  \bibinfo{author}{I.~Serban}, \bibinfo{author}{J.~Pineau},
\newblock \bibinfo{title}{The ubuntu dialogue corpus: A large dataset for
  research in unstructured multi-turn dialogue systems},
\newblock in: \bibinfo{booktitle}{Proceedings of the 16th Annual Meeting of the
  Special Interest Group on Discourse and Dialogue}, \bibinfo{year}{2015}, pp.
  \bibinfo{pages}{285--294}.
\bibitem[{Park et~al.(2018)Park, Cho, and Kim}]{park2018hierarchical}
\bibinfo{author}{Y.~Park}, \bibinfo{author}{J.~Cho}, \bibinfo{author}{G.~Kim},
\newblock \bibinfo{title}{A hierarchical latent structure for variational
  conversation modeling},
\newblock in: \bibinfo{booktitle}{Proceedings of the 2018 Conference of the
  North American Chapter of the Association for Computational Linguistics:
  Human Language Technologies, Volume 1 (Long Papers)}, \bibinfo{year}{2018},
  pp. \bibinfo{pages}{1792--1801}.
\bibitem[{Busso et~al.(2008)Busso, Bulut, Lee, Kazemzadeh, Mower, Kim, Chang,
  Lee, and Narayanan}]{busso2008iemocap}
\bibinfo{author}{C.~Busso}, \bibinfo{author}{M.~Bulut}, \bibinfo{author}{C.-C.
  Lee}, \bibinfo{author}{A.~Kazemzadeh}, \bibinfo{author}{E.~Mower},
  \bibinfo{author}{S.~Kim}, \bibinfo{author}{J.~N. Chang},
  \bibinfo{author}{S.~Lee}, \bibinfo{author}{S.~S. Narayanan},
\newblock \bibinfo{title}{Iemocap: Interactive emotional dyadic motion capture
  database},
\newblock \bibinfo{journal}{Language resources and evaluation}
  \bibinfo{volume}{42} (\bibinfo{year}{2008}) \bibinfo{pages}{335}.
\bibitem[{Li et~al.(2017)Li, Su, Shen, Li, Cao, and Niu}]{li2017dailydialog}
\bibinfo{author}{Y.~Li}, \bibinfo{author}{H.~Su}, \bibinfo{author}{X.~Shen},
  \bibinfo{author}{W.~Li}, \bibinfo{author}{Z.~Cao}, \bibinfo{author}{S.~Niu},
\newblock \bibinfo{title}{Dailydialog: A manually labelled multi-turn dialogue
  dataset},
\newblock in: \bibinfo{booktitle}{Proceedings of the Eighth International Joint
  Conference on Natural Language Processing (Volume 1: Long Papers)},
  \bibinfo{year}{2017}, pp. \bibinfo{pages}{986--995}.
\bibitem[{Schuller et~al.(2012)Schuller, Valstar, Eyben, Cowie, and
  Pantic}]{schuller2012avec}
\bibinfo{author}{B.~W. Schuller}, \bibinfo{author}{M.~F. Valstar},
  \bibinfo{author}{F.~Eyben}, \bibinfo{author}{R.~Cowie},
  \bibinfo{author}{M.~Pantic},
\newblock \bibinfo{title}{{AVEC} 2012: the continuous audio/visual emotion
  challenge},
\newblock in: \bibinfo{editor}{L.~Morency}, \bibinfo{editor}{D.~Bohus},
  \bibinfo{editor}{H.~K. Aghajan}, \bibinfo{editor}{J.~Cassell},
  \bibinfo{editor}{A.~Nijholt}, \bibinfo{editor}{J.~Epps} (Eds.),
  \bibinfo{booktitle}{International Conference on Multimodal Interaction,
  {ICMI} '12, Santa Monica, CA, USA, October 22-26, 2012},
  \bibinfo{publisher}{{ACM}}, \bibinfo{year}{2012}, pp.
  \bibinfo{pages}{449--456}. \URLprefix
  \url{https://doi.org/10.1145/2388676.2388776}.
  \DOIprefix\doi{10.1145/2388676.2388776}.
\bibitem[{Kim(2014)}]{DBLP:conf/emnlp/Kim14}
\bibinfo{author}{Y.~Kim},
\newblock \bibinfo{title}{Convolutional neural networks for sentence
  classification},
\newblock in: \bibinfo{editor}{A.~Moschitti}, \bibinfo{editor}{B.~Pang},
  \bibinfo{editor}{W.~Daelemans} (Eds.), \bibinfo{booktitle}{Proceedings of the
  2014 Conference on Empirical Methods in Natural Language Processing, {EMNLP}
  2014, October 25-29, 2014, Doha, Qatar, {A} meeting of SIGDAT, a Special
  Interest Group of the {ACL}}, \bibinfo{publisher}{{ACL}},
  \bibinfo{year}{2014}, pp. \bibinfo{pages}{1746--1751}. \URLprefix
  \url{https://doi.org/10.3115/v1/d14-1181}.
  \DOIprefix\doi{10.3115/v1/d14-1181}.
\bibitem[{Sukhbaatar et~al.(2015)Sukhbaatar, Szlam, Weston, and
  Fergus}]{DBLP:conf/nips/SukhbaatarSWF15}
\bibinfo{author}{S.~Sukhbaatar}, \bibinfo{author}{A.~Szlam},
  \bibinfo{author}{J.~Weston}, \bibinfo{author}{R.~Fergus},
\newblock \bibinfo{title}{End-to-end memory networks},
\newblock in: \bibinfo{editor}{C.~Cortes}, \bibinfo{editor}{N.~D. Lawrence},
  \bibinfo{editor}{D.~D. Lee}, \bibinfo{editor}{M.~Sugiyama},
  \bibinfo{editor}{R.~Garnett} (Eds.), \bibinfo{booktitle}{Advances in Neural
  Information Processing Systems 28: Annual Conference on Neural Information
  Processing Systems 2015, December 7-12, 2015, Montreal, Quebec, Canada},
  \bibinfo{year}{2015}, pp. \bibinfo{pages}{2440--2448}. \URLprefix
  \url{http://papers.nips.cc/paper/5846-end-to-end-memory-networks}.
\bibitem[{Hochreiter and Schmidhuber(1997)}]{DBLP:journals/neco/HochreiterS97}
\bibinfo{author}{S.~Hochreiter}, \bibinfo{author}{J.~Schmidhuber},
\newblock \bibinfo{title}{Long short-term memory},
\newblock \bibinfo{journal}{Neural Computation} \bibinfo{volume}{9}
  (\bibinfo{year}{1997}) \bibinfo{pages}{1735--1780}. \URLprefix
  \url{https://doi.org/10.1162/neco.1997.9.8.1735}.
  \DOIprefix\doi{10.1162/neco.1997.9.8.1735}.
\bibitem[{Poria et~al.(2017)Poria, Cambria, Hazarika, Majumder, Zadeh, and
  Morency}]{DBLP:conf/icdm/PoriaCHMZM17}
\bibinfo{author}{S.~Poria}, \bibinfo{author}{E.~Cambria},
  \bibinfo{author}{D.~Hazarika}, \bibinfo{author}{N.~Majumder},
  \bibinfo{author}{A.~Zadeh}, \bibinfo{author}{L.~Morency},
\newblock \bibinfo{title}{Multi-level multiple attentions for contextual
  multimodal sentiment analysis},
\newblock in: \bibinfo{editor}{V.~Raghavan}, \bibinfo{editor}{S.~Aluru},
  \bibinfo{editor}{G.~Karypis}, \bibinfo{editor}{L.~Miele},
  \bibinfo{editor}{X.~Wu} (Eds.), \bibinfo{booktitle}{2017 {IEEE} International
  Conference on Data Mining, {ICDM} 2017, New Orleans, LA, USA, November 18-21,
  2017}, \bibinfo{publisher}{{IEEE} Computer Society}, \bibinfo{year}{2017},
  pp. \bibinfo{pages}{1033--1038}. \URLprefix
  \url{https://doi.org/10.1109/ICDM.2017.134}.
  \DOIprefix\doi{10.1109/ICDM.2017.134}.
\bibitem[{Tieleman and Hinton(2012)}]{Tieleman2012}
\bibinfo{author}{T.~Tieleman}, \bibinfo{author}{G.~Hinton},
  \bibinfo{title}{{Lecture 6.5---RmsProp: Divide the gradient by a running
  average of its recent magnitude}}, \bibinfo{howpublished}{COURSERA: Neural
  Networks for Machine Learning}, \bibinfo{year}{2012}.
\bibitem[{Kingma and Ba(2015)}]{kingma2014adam}
\bibinfo{author}{D.~P. Kingma}, \bibinfo{author}{J.~Ba},
\newblock \bibinfo{title}{Adam: {A} method for stochastic optimization},
\newblock in: \bibinfo{editor}{Y.~Bengio}, \bibinfo{editor}{Y.~LeCun} (Eds.),
  \bibinfo{booktitle}{3rd International Conference on Learning Representations,
  {ICLR} 2015, San Diego, CA, USA, May 7-9, 2015, Conference Track
  Proceedings}, \bibinfo{year}{2015}. \URLprefix
  \url{http://arxiv.org/abs/1412.6980}.
\bibitem[{Nachar et~al.(2008)}]{nachar2008mann}
\bibinfo{author}{N.~Nachar}, et~al.,
\newblock \bibinfo{title}{The mann-whitney u: A test for assessing whether two
  independent samples come from the same distribution},
\newblock \bibinfo{journal}{Tutorials in quantitative Methods for Psychology}
  \bibinfo{volume}{4} (\bibinfo{year}{2008}) \bibinfo{pages}{13--20}.
\bibitem[{Rajpurkar et~al.(2018)Rajpurkar, Jia, and
  Liang}]{DBLP:conf/acl/RajpurkarJL18}
\bibinfo{author}{P.~Rajpurkar}, \bibinfo{author}{R.~Jia},
  \bibinfo{author}{P.~Liang},
\newblock \bibinfo{title}{Know what you don't know: Unanswerable questions for
  squad},
\newblock in: \bibinfo{editor}{I.~Gurevych}, \bibinfo{editor}{Y.~Miyao} (Eds.),
  \bibinfo{booktitle}{Proceedings of the 56th Annual Meeting of the Association
  for Computational Linguistics, {ACL} 2018, Melbourne, Australia, July 15-20,
  2018, Volume 2: Short Papers}, \bibinfo{publisher}{Association for
  Computational Linguistics}, \bibinfo{year}{2018}, pp.
  \bibinfo{pages}{784--789}. \URLprefix
  \url{https://www.aclweb.org/anthology/P18-2124/}.
  \DOIprefix\doi{10.18653/v1/P18-2124}.
\bibitem[{Williams et~al.(2018)Williams, Nangia, and Bowman}]{N18-1101}
\bibinfo{author}{A.~Williams}, \bibinfo{author}{N.~Nangia},
  \bibinfo{author}{S.~Bowman},
\newblock \bibinfo{title}{A broad-coverage challenge corpus for sentence
  understanding through inference},
\newblock in: \bibinfo{booktitle}{Proceedings of the 2018 Conference of the
  North American Chapter of the Association for Computational Linguistics:
  Human Language Technologies, Volume 1 (Long Papers)},
  \bibinfo{publisher}{Association for Computational Linguistics},
  \bibinfo{year}{2018}, pp. \bibinfo{pages}{1112--1122}. \URLprefix
  \url{http://aclweb.org/anthology/N18-1101}.
\bibitem[{Mohammad and Turney(2013)}]{mohammad2013crowdsourcing}
\bibinfo{author}{S.~Mohammad}, \bibinfo{author}{P.~D. Turney},
\newblock \bibinfo{title}{Crowdsourcing a word-emotion association lexicon},
\newblock \bibinfo{journal}{Computational Intelligence} \bibinfo{volume}{29}
  (\bibinfo{year}{2013}) \bibinfo{pages}{436--465}. \URLprefix
  \url{https://doi.org/10.1111/j.1467-8640.2012.00460.x}.
  \DOIprefix\doi{10.1111/j.1467-8640.2012.00460.x}.
\bibitem[{Peters et~al.(2019)Peters, Ruder, and Smith}]{peters2019tune}
\bibinfo{author}{M.~E. Peters}, \bibinfo{author}{S.~Ruder},
  \bibinfo{author}{N.~A. Smith},
\newblock \bibinfo{title}{To tune or not to tune? adapting pretrained
  representations to diverse tasks},
\newblock in: \bibinfo{editor}{I.~Augenstein}, \bibinfo{editor}{S.~Gella},
  \bibinfo{editor}{S.~Ruder}, \bibinfo{editor}{K.~Kann},
  \bibinfo{editor}{B.~Can}, \bibinfo{editor}{J.~Welbl},
  \bibinfo{editor}{A.~Conneau}, \bibinfo{editor}{X.~Ren},
  \bibinfo{editor}{M.~Rei} (Eds.), \bibinfo{booktitle}{Proceedings of the 4th
  Workshop on Representation Learning for NLP, RepL4NLP@ACL 2019, Florence,
  Italy, August 2, 2019.}, \bibinfo{publisher}{Association for Computational
  Linguistics}, \bibinfo{year}{2019}, pp. \bibinfo{pages}{7--14}. \URLprefix
  \url{https://www.aclweb.org/anthology/W19-4302/}.
\bibitem[{Serban et~al.(2017)Serban, Sordoni, Lowe, Charlin, Pineau, Courville,
  and Bengio}]{SerbanSLCPCB17}
\bibinfo{author}{I.~V. Serban}, \bibinfo{author}{A.~Sordoni},
  \bibinfo{author}{R.~Lowe}, \bibinfo{author}{L.~Charlin},
  \bibinfo{author}{J.~Pineau}, \bibinfo{author}{A.~C. Courville},
  \bibinfo{author}{Y.~Bengio},
\newblock \bibinfo{title}{A hierarchical latent variable encoder-decoder model
  for generating dialogues},
\newblock in: \bibinfo{booktitle}{Proceedings of the Thirty-First {AAAI}
  Conference on Artificial Intelligence, February 4-9, 2017, San Francisco,
  California, {USA.}}, \bibinfo{year}{2017}, pp. \bibinfo{pages}{3295--3301}.
  \URLprefix \url{http://aaai.org/ocs/index.php/AAAI/AAAI17/paper/view/14567}.
\bibitem[{Mower et~al.(2011)Mower, Mataric, and
  Narayanan}]{DBLP:journals/taslp/MowerMN11}
\bibinfo{author}{E.~Mower}, \bibinfo{author}{M.~J. Mataric},
  \bibinfo{author}{S.~S. Narayanan},
\newblock \bibinfo{title}{A framework for automatic human emotion
  classification using emotion profiles},
\newblock \bibinfo{journal}{{IEEE} Trans. Audio, Speech {\&} Language
  Processing} \bibinfo{volume}{19} (\bibinfo{year}{2011})
  \bibinfo{pages}{1057--1070}. \URLprefix
  \url{https://doi.org/10.1109/TASL.2010.2076804}.
  \DOIprefix\doi{10.1109/TASL.2010.2076804}.
\bibitem[{Li et~al.(2016)Li, Galley, Brockett, Gao, and
  Dolan}]{li-etal-2016-diversity}
\bibinfo{author}{J.~Li}, \bibinfo{author}{M.~Galley},
  \bibinfo{author}{C.~Brockett}, \bibinfo{author}{J.~Gao},
  \bibinfo{author}{B.~Dolan},
\newblock \bibinfo{title}{A diversity-promoting objective function for neural
  conversation models},
\newblock in: \bibinfo{booktitle}{Proceedings of the 2016 Conference of the
  North {A}merican Chapter of the Association for Computational Linguistics:
  Human Language Technologies}, \bibinfo{publisher}{Association for
  Computational Linguistics}, \bibinfo{address}{San Diego, California},
  \bibinfo{year}{2016}, pp. \bibinfo{pages}{110--119}. \URLprefix
  \url{https://www.aclweb.org/anthology/N16-1014}.
  \DOIprefix\doi{10.18653/v1/N16-1014}.
\bibitem[{Song et~al.(2018)Song, Li, Nie, Zhang, Zhao, and
  Yan}]{DBLP:conf/ijcai/SongLNZZY18}
\bibinfo{author}{Y.~Song}, \bibinfo{author}{C.~Li}, \bibinfo{author}{J.~Nie},
  \bibinfo{author}{M.~Zhang}, \bibinfo{author}{D.~Zhao},
  \bibinfo{author}{R.~Yan},
\newblock \bibinfo{title}{An ensemble of retrieval-based and generation-based
  human-computer conversation systems},
\newblock in: \bibinfo{booktitle}{Proceedings of the Twenty-Seventh
  International Joint Conference on Artificial Intelligence, {IJCAI} 2018, July
  13-19, 2018, Stockholm, Sweden.}, \bibinfo{year}{2018}, pp.
  \bibinfo{pages}{4382--4388}. \URLprefix
  \url{https://doi.org/10.24963/ijcai.2018/609}.
  \DOIprefix\doi{10.24963/ijcai.2018/609}.
\bibitem[{Zhou et~al.(2018)Zhou, Huang, Zhang, Zhu, and
  Liu}]{DBLP:conf/aaai/ZhouHZZL18}
\bibinfo{author}{H.~Zhou}, \bibinfo{author}{M.~Huang},
  \bibinfo{author}{T.~Zhang}, \bibinfo{author}{X.~Zhu},
  \bibinfo{author}{B.~Liu},
\newblock \bibinfo{title}{Emotional chatting machine: Emotional conversation
  generation with internal and external memory},
\newblock in: \bibinfo{booktitle}{Proceedings of the Thirty-Second {AAAI}
  Conference on Artificial Intelligence, (AAAI-18), the 30th innovative
  Applications of Artificial Intelligence (IAAI-18), and the 8th {AAAI}
  Symposium on Educational Advances in Artificial Intelligence (EAAI-18), New
  Orleans, Louisiana, USA, February 2-7, 2018}, \bibinfo{year}{2018}, pp.
  \bibinfo{pages}{730--739}. \URLprefix
  \url{https://www.aaai.org/ocs/index.php/AAAI/AAAI18/paper/view/16455}.
\bibitem[{Burstein et~al.(2019)Burstein, Doran, and
  Solorio}]{DBLP:conf/naacl/2019-1}
\bibinfo{editor}{J.~Burstein}, \bibinfo{editor}{C.~Doran},
  \bibinfo{editor}{T.~Solorio} (Eds.), \bibinfo{title}{Proceedings of the 2019
  Conference of the North American Chapter of the Association for Computational
  Linguistics: Human Language Technologies, {NAACL-HLT} 2019, Minneapolis, MN,
  USA, June 2-7, 2019, Volume 1 (Long and Short Papers)},
  \bibinfo{publisher}{Association for Computational Linguistics},
  \bibinfo{year}{2019}. \URLprefix
  \url{https://www.aclweb.org/anthology/volumes/N19-1/}.
\bibitem[{Riloff et~al.(2018)Riloff, Chiang, Hockenmaier, and
  Tsujii}]{DBLP:conf/emnlp/2018}
\bibinfo{editor}{E.~Riloff}, \bibinfo{editor}{D.~Chiang},
  \bibinfo{editor}{J.~Hockenmaier}, \bibinfo{editor}{J.~Tsujii} (Eds.),
  \bibinfo{title}{Proceedings of the 2018 Conference on Empirical Methods in
  Natural Language Processing, Brussels, Belgium, October 31 - November 4,
  2018}, \bibinfo{publisher}{Association for Computational Linguistics},
  \bibinfo{year}{2018}. \URLprefix
  \url{https://www.aclweb.org/anthology/volumes/D18-1/}.
\bibitem[{Kraus(2019)}]{DBLP:conf/ijcai/2019}
\bibinfo{editor}{S.~Kraus} (Ed.), \bibinfo{title}{Proceedings of the
  Twenty-Eighth International Joint Conference on Artificial Intelligence,
  {IJCAI} 2019, Macao, China, August 10-16, 2019},
  \bibinfo{publisher}{ijcai.org}, \bibinfo{year}{2019}. \URLprefix
  \url{https://doi.org/10.24963/ijcai.2019}.
  \DOIprefix\doi{10.24963/ijcai.2019}.

\end{thebibliography}

\end{document}